\documentclass[final,5p,times,twocolumn,authoryear]{elsarticle}
\usepackage{amsfonts}
\usepackage{graphicx}
\usepackage{soul}
\usepackage{textcomp}
\usepackage{setspace}
\usepackage{placeins}
\usepackage[utf8]{inputenc} 
\usepackage{tabularx}
\usepackage{multirow}
\usepackage{adjustbox}
\usepackage{tabulary}
\usepackage{textcomp}
\usepackage{mathtools}
\usepackage{mwe}
\usepackage[noend]{algpseudocode}
\usepackage{float}
\usepackage{parskip}
\usepackage{amsmath}
\usepackage{mathrsfs}
\usepackage{amsthm}
\usepackage{hyperref}
\usepackage{amssymb}
\usepackage{tabularx}
\usepackage{caption}
\usepackage{subcaption}
\usepackage{gensymb}
\usepackage{bm}
\usepackage{geometry}
\usepackage{algorithm}
\usepackage[capitalise, nameinlink]{cleveref}

\usepackage[percent]{overpic}

\usepackage{color,soul}

\DeclarePairedDelimiter\floor{\lfloor}{\rfloor}

\DeclareMathOperator*{\argmin}{arg\,min}
\DeclareMathOperator{\atantwo}{atan2}
\newcommand{\integerset}[1]{\left\{1,\ldots,{#1}\right\}}

\usepackage{array}
\newcolumntype{L}{>{\arraybackslash}p{5cm}}
\newlength{\temp}
\newtheorem{assumption}{Assumption}

\newgeometry{vmargin={20mm}, hmargin={20mm,20mm}}   

\usepackage{comment}
\begin{document}
\begin{frontmatter}
\title{Risk-based implementation of COLREGs for autonomous surface vehicles using deep reinforcement learning}
\author[adilthomassaddress]{Thomas Nakken Larsen}

\author[amaliesaddress]{Amalie Heiberg}

\author[eivindaddress]{Eivind Meyer}

\author[adilthomassaddress]{Adil Rasheed\corref{mycorrespondingauthor}}
\cortext[mycorrespondingauthor]{Adil Rasheed}
\ead{adil.rasheed@ntnu.no}

\author[omersaddress]{Omer San}

\author[adilthomassaddress]{Damiano Varagnolo}
\address[adilthomassaddress]{Department of Engineering Cybernetics, Norwegian University of Science and Technology}
\address[amaliesaddress]{Equinor}
\address[eivindaddress]{Institute of Informatics, Technical University of Munich}
\address[omersaddress]{School of Mechanical and Aerospace Engineering, Oklahoma State University}

\begin{abstract}
Autonomous systems are becoming ubiquitous and gaining momentum within the marine sector. Since the electrification of transport is happening simultaneously, autonomous marine vessels can reduce environmental impact, lower costs, and increase efficiency. Although close monitoring is still required to ensure safety, the ultimate goal is full autonomy. One major milestone is to develop a control system that is versatile enough to handle any weather and encounter that is also robust and reliable. Additionally, the control system must adhere to the International Regulations for Preventing Collisions at Sea (COLREGs) for successful interaction with human sailors. Since the COLREGs were written for the human mind to interpret, they are written in ambiguous prose and therefore not machine-readable or verifiable. Due to these challenges and the wide variety of situations to be tackled, classical model-based approaches prove complicated to implement and computationally heavy. Within machine learning (ML), deep reinforcement learning (DRL) has shown great potential for a wide range of applications. The model-free and self-learning properties of DRL make it a promising candidate for autonomous vessels. In this work, a subset of the COLREGs is incorporated into a DRL-based path following and obstacle avoidance system using collision risk theory. The resulting autonomous agent dynamically interpolates between path following and COLREG-compliant collision avoidance in the training scenario, isolated encounter situations, and AIS-based simulations of real-world scenarios. 
\end{abstract}




\end{frontmatter}
\section{Introduction}
\label{sec:Introduction}
Over the last few years, the promising idea of autonomous ships has gained traction through projects like ReVolt (\cite{ReVolt}) and Yara Birkeland (\cite{YaraBirkeland}). Such research projects are increasingly incentivized as the funding bodies recognize the potential benefits of autonomy at sea. A notable example is the EU-funded four-year project Autoship Horizon 2020, which seeks to speed up the transition towards autonomous ships in the EU (\cite{Autoship}). For the first time in history, the promise of lower emissions, higher efficiency, and fewer accidents via autonomy is becoming tangible.

Human error is a leading cause of accidents on the road (\cite{Dingus2016, Thomas2013}), and reports show that accidents at sea are no different. According to the Annual Overview of Marine Casualties and Incidents published by the European Maritime Safety Agency (EMSA), human error was attributed to over 50\% of accidental events between 2011-17 (\cite{EMSA}). In addition to reducing accidents (and thereby fatalities), environmental damage, and costs, autonomous marine operations allow for optimized route planning. This optimization can be done with respect to time spent and fuel costs. Furthermore, autonomous ships can move cargo transport from the road to the sea, leading to less trafficked roads. For instance, the autonomous container ship Yara Birkeland is expected to reduce the number of trips made by diesel trucks by 40,000 a year after its launch in 2020 (\cite{YaraZeroEmission}). With the widespread electrification taking place, reduced air pollution is another likely and desirable effect. 

An overall reduction of errors from introducing autonomy depends on developing robust and reliable systems, which is no trivial task. For autonomous navigation at sea, the vessel's control system must deal appropriately with a wide range of situations depending on the position of the ownship (OS) and other ships within a certain radius and environmental factors such as wind and ocean currents, and waves. Another crucial element is detection, classification, and tracking of objects, which might be challenging in certain weather conditions. The currently proposed solutions generally make significant simplifications and assumptions. Low-level controllers, or autopilots, are already commercially available, but more research on high-level path planning and collision avoidance is needed to ensure safe autonomous navigation in real situations. For collision avoidance, compliance with the International Regulations for Preventing Collisions at Sea (COLREGs) is crucial to ensure safety when encountering other vessels.

Due to the complex nature of autonomy at sea, classical model-based methods may be challenging to implement for full autonomy. Modern machine learning (ML) methods are proficient in approximating such complex models. Supervised learning approaches are powerful but limited by their dependency on labeled training data. Reinforcement learning (RL) circumvents this by producing the training data as the decision-making agent interacts with its environment. However, there exists limited research on the combined topic of COLREG-compliant RL controllers. Therefore, this work aims to incorporate a subset of the COLREGs, directly related to collision avoidance, into an autonomous path following and collision avoidance system based on deep reinforcement learning (DRL) conditioned on measures of collision risk. Thus, the research questions are as follows:
\begin{itemize}
    \item Can we use state-of-the-art collision risk theory to define a novel risk-based approach for implementing COLREGs into an autonomous DRL agent?
    \item Can the risk-based DRL agent learn to intelligently interpolate between path following and collision avoidance while maneuvering in a COLREG-compliant manner?
\end{itemize}

\Cref{sec:background} describes the state-of-the-art in collision avoidance for marine guidance, in which the COLREGs are generally ignored. \Cref{sec:theory} introduces the COLREGs relevant for collision avoidance and essential concepts within guidance and control, collision risk theory, and DRL. \Cref{section:methodology} defines the simulation environments, DRL problem formulation, and evaluation methods. \Cref{sec:results} presents and discusses the COLREG-compliance and general path following and collision avoidance performance of the resulting autonomous controller. \Cref{sec:conclusion} summarizes the findings and suggests future work.

\section{Background}
\label{sec:background}
Collision alert systems (CAS) aid the captain and crew on board a marine vessel. Such systems primarily extend exteroceptive sensors, converting raw measurements into more interpretable information. Examples of CAS systems are Automatic Radar Plotting Aid (ARPA) and Automatic Identification System (AIS) (compared in \cite{Lin2006}), routinely used for collision risk evaluation (\cite{Xu2014}). As we move into the fourth industrial revolution, solutions such as digital twins and remote sensing are making their way into the maritime industry (\cite{DigTwin}). Decision-making is thus gradually being taken from the cognitive realm and into the digital domain, and the need for highly robust and flexible guidance, navigation, and control (GNC) systems is growing. Since collision avoidance (COLAV) systems are responsible for one of the most safety-critical aspects of a vessel's operation, any GNC system operating in a dynamic environment requires a robust COLAV strategy (\cite{Aniculaesei2016}). Therefore, reliable and transparent COLAV systems are crucial to reach full autonomy at sea. 

All vessels above 300 tonnes engaged on international voyages, all cargo ships above 500 tonnes, and all passenger ships are required to carry an AIS (\cite{AIS}). The AIS transmits and receives information such as identity, position, course, and speed, which can be incorporated into a COLAV system. Such systems can thus enhance the quality of information about other vessels but may also depend on the communication infrastructure. Since one cannot expect complete availability, ships typically utilize additional exteroceptive sensors such as cameras, lidars, and radars. Ideally, an autonomous COLAV system uses AIS information without depending on it.

Before autonomous vessels became a possibility, the International Regulations for Preventing Collisions at Sea (COLREGs) were formulated to prevent collisions between two or more vessels (\cite{COLREGs}). Although technological advancement has been significant since their publication in 1972, COLREG-compliance for autonomous vessels is still understudied. One of the main challenges is that the COLREGs were written for humans to interpret and require a translation to a machine-readable and verifiable format. Another potential challenge is the indirect communication that occurs when two vessels meet in a situation with a high risk of collision. For instance, the COLREGs require sharp maneuvers for clear communication between vessels when a high-risk situation is encountered. However, this is often not the optimal behavior from an energy efficiency (or even collision risk) point of view. So long as there may be both human and autonomous operators of marine vessels at sea, the autonomous controller should behave in a way that a human-operated vessel can interpret its intent.

In addition to the challenges inherent to the COLREGs, autonomous collision avoidance can be demanding due to the complex dynamics of ships, varying speeds, and changing environmental conditions (\cite{Tam2009}). The majority of the proposed solutions for autonomy make assumptions that do not represent reality. Examples of such assumptions are the constant speed of the OS or other ships, good weather conditions, or that the system only operates while the ship is at open sea. An adequate autonomous vessel must master all the situations the current fleet handles. For instance, given sufficient situational awareness, a full-fledged autonomous COLAV system should be expected to handle situations involving all sorts of moving and stationary objects, from container ships to kayaks. For generalization, the system must track a high number of objects simultaneously and perform well in congested waters.

A plethora of COLAV algorithms and architectures for autonomous control have been, and still are, researched. Here, we distinguish between \textit{classical} and \textit{soft} systems (\cite{Statheros2008}). Classical systems find an optimal strategy analytically and numerically from mathematical models and logic, which are typically accompanied by convergence proofs. Model predictive control (MPC) can be used to develop COLAV systems compliant with the primary rules of COLREGs. MPC can also be applied to nonlinear systems with uncertain environmental disturbances (\cite{Soloperto2019}). The Velocity Obstacle (VO) method (\cite{Fiorini1998}) models artificial obstacles representing the velocities that would result in a collision, and \cite{Kuwata2014} shows that maritime navigation using the VO method can be COLREGs-compliant. Interval Programming (IvP), a multi-objective optimization approach, has successfully produced COLREGs-compliant COLAV systems (\cite{Benjamin2006, Woernner2014}). Dynamic Window (DW) is an optimization-based method that has been researched for marine applications (\cite{Serigstad2018}), the strength of which can be found in its focus on fast dynamics through reducing the search space to the reachable velocities within a short time interval \cite{Fox2016}. 

Based on artificial intelligence (AI), \textit{soft systems} assume that the problem is not readily quantified. Heuristics are experience-based methods for finding an acceptable solution to a problem. The A* heuristic (\cite{Hart1968}) might be the most well-known and widely used soft approach; A* is a greedy search algorithm for finding the shortest distance between two nodes in a graph, in which a heuristic measure weights the edges between nodes. It is often used for high-level path and trajectory planning, as was done in \cite{Eriksen2019}. Another well-known heuristic is the genetic algorithm (GA) based on evolutionary theory. \cite{Smierzchalski1999} applies a genetic algorithm for trajectory planning in an environment with static and dynamic obstacles. \cite{Kim2015} showed that Distributed Tabu Search, a metaheuristic method, can be used for collision avoidance in highly congested areas. Another group of soft systems is machine learning (ML). ML techniques such as deep learning (DL) and reinforcement learning (RL) have recently gotten significant attention in the context of autonomous systems and decision-making problems, as they benefit from neural networks' currently unmatched function approximation capabilities. Model-free RL methods can find a control law even without any mathematical model of the system (\cite{SILVER2021RewardIsEnough}). However, only a limited amount of research has been devoted to autonomous marine vessels compared to driver-less cars, for instance. In \cite{Xu2017}, a deep convolutional neural network (CNN) is trained for COLREGs-compliant collision avoidance for a crewless surface vehicle. This method is based on image recognition, using the CNN's ability to process spatially structured data. The Deep Deterministic Policy Gradient (DDPG) algorithm has demonstrated successful path following and simple collision avoidance for marine vessel models (\cite{Martinsen2018, Martinsen2019, Vallestad2029}). In addition, \cite{Zhao2019a, meyer_ASV_IEEE} showcased the Proximal Policy Optimization (PPO) algorithm for multi-ship collision avoidance.  

Alternatively, COLAV systems can be classified as \textit{deliberative} or \textit{reactive} systems (\cite{delib_react}). Deliberative systems work in a ``sense-plan-act'' fashion. Intuitively, reactive systems are then considered ``sense-act'' systems.
Hybrid COLAV systems emerge when combining different system categories, e.g., deliberate and reactive systems. This approach is made with increasing frequency (\cite{Ding2011}). Multi-layered systems are also being developed, where each subsystem lies on a spectrum between reactive and deliberative. Such hybrid architectures are able to harvest the strengths of several methods, using each where they perform best. \cite{Loe2008} applies a two-layered approach where deliberation is done by a Rapidly-Exploring Random Tree (RRT) algorithm combined with the deliberative A* heuristic, and the reactive component consists of a modified DW algorithm. In \cite{Eriksen2019}, A* is combined with a mid-layer and a reactive MPC-based algorithm, forming a three-layered COLAV system. \cite{Casalino2009,Svec2013} have proposed similar layered architectures.

In summary, a wide range of COLAV systems have been proposed in literature, generally disregarding the COLREGs. At the same time, the increased focus on autonomous systems in later years requires COLREG-compliance for sufficient safety. This gap combined with the promise of DRL for autonomous navigation, shapes the objective of the article --- to investigate COLREG-compliance in a path following and collision avoidance system based on deep reinforcement learning, conditioned on measures of collision risk.

\section{Theory}\label{sec:theory}
\subsection{Dynamics of a marine vessel}
\begin{figure}[ht!]
    \centering
    \includegraphics[width=0.8\linewidth]{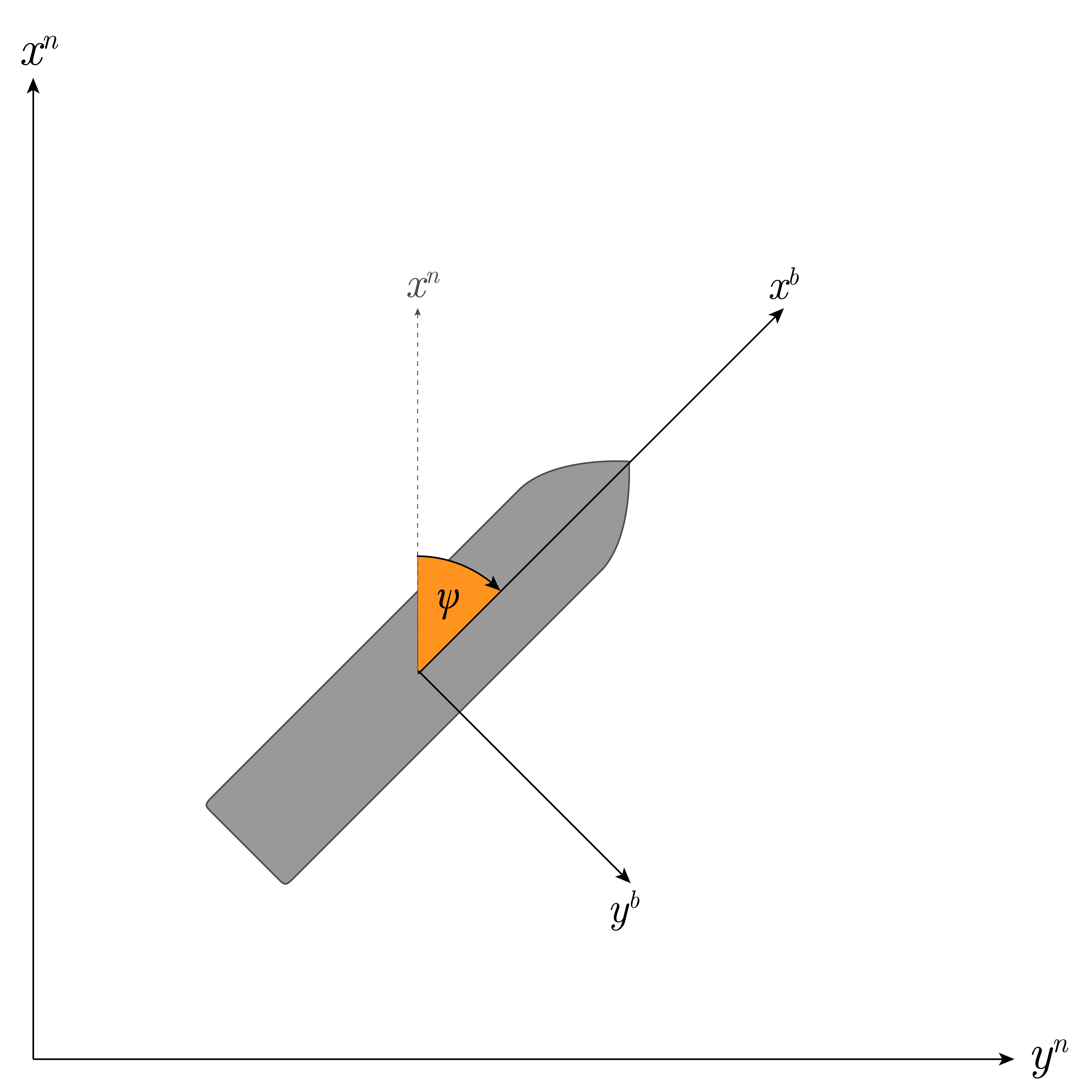}
	\caption[Coordinate frames]{Illustration of the NED and body coordinate frames.}
	\label{fig:coordinateframes}
\end{figure}

The dynamical model considered in this work is CyberShip II: a 1:70 scale replica of a supply ship (\cite{Skjetne2004ANS}). This model is simulated in a calm ocean surface environment with the following assumptions.
\begin{assumption}[State space restriction]\label{as:restricted_motion}The vessel is always located on the surface, and thus there is no heave motion. Also, there is no pitching or rolling motion.\end{assumption}
\begin{assumption}[Calm sea]\label{as:calm_sea} There are no external disturbances to the vessel, such as wind, ocean currents, or waves.\end{assumption}
Following SNAME notation (\cite{SNAME}), the navigation state vector then consists of the generalized coordinates, $\bm{\eta} = \left[ x^n, y^n, \psi \right]^T$, where $x^n$ and $y^n$ are the North and East positions, respectively, in the North-East-Down (NED) reference frame $\{n\}$, and $\psi$ is the yaw angle, i.e., the current angle between the vessel's longitudinal axis $x_b$ and the North axis $x_n$, illustrated by \Cref{fig:coordinateframes}. Correspondingly, the translational and angular velocity vector $\bm{\nu} = \left[ u, v, r \right]^T$ consists of the surge (i.e., forward) velocity $u$, the sway (i.e., sideways) velocity $v$ and yaw rate $r$.

\subsubsection{Vessel model}\label{section:vessel_dynamics}
Given the established assumptions, the 3-DOF vessel dynamics can be expressed in a compact matrix-vector form 

\begin{equation*}
    \begin{aligned}
        \dot{\bm{\eta}} &= \mathbf{R}_{z,\psi}(\bm{\eta})   \bm{\nu} \\
        \mathbf{M} \bm{\dot{\nu}} + \mathbf{C}(\bm{\nu}) \bm{\nu} + \mathbf{D}(\bm{\nu}) \bm{\nu} &= \mathbf{B} \bm{f},
    \end{aligned}
\end{equation*}
where $\mathbf{R}_{z,\psi}$ represents a rotation of $\psi$ radians around the $z_n$-axis as defined by
\begin{equation*}
    \begin{aligned}
        \mathbf{R}_{z,\psi} &= \begin{bmatrix}
                                    \cos{\psi} & -\sin{\psi} & 0 \\
                                    \sin{\psi} & \cos{\psi} & 0 \\
                                    0 & 0 & 1
                                \end{bmatrix}
    \end{aligned}
\end{equation*}
\noindent Furthermore, $\mathbf{M} \in \mathbb{R}^{3\times3}$ is the mass matrix and includes the effects of both rigid-body and added mass, $\mathbf{C}(\bm{\nu}) \in \mathbb{R}^{3\times3}$ incorporates centripetal and Coriolis effects, and $\mathbf{D}(\bm{\nu}) \in \mathbb{R}^{3\times3}$ is the damping matrix. Finally, $\mathbf{B} \in \mathbb{R}^{3\times2}$ is the actuator configuration matrix. The numerical values of the matrices are found in \cite{SKJETNE2004203}, where the model parameters were estimated experimentally for CyberShip II in a marine control laboratory.

We disregard the ship's bow thruster and allow only the aft thrusters and control surfaces to be applied by the Reinforcement Learning (RL) agent as control signals. This omission simplifies the RL agent's action space and is further motivated by the bow thrusters' limited effectiveness at higher speeds (\cite{ShipHeadingSpeedControlSorensenBreivikEriksen}). Thus, the control vector, $\bm{f} = \left[T_u, T_r\right]^T$, consists of the surge force input, $T_u$, and the yaw's moment input, $T_r$.

\subsubsection{COLREG rules}\label{sec:relevant_rules}

Among the 41 rules in the International Regulations for Preventing Collisions at Sea (\cite{COLREGsPDF}), only the directly relevant rules for COLAV are presented below. The two main takeaways from these rules are that 1) the give-way vessel should take early and substantial action, and 2) safe speed should be ensured at all times, such that course alteration is effective towards avoiding collisions where there is sufficient sea-room. Since rules 6 and 8 are particularly tough to quantify, this work focuses on compliance to rules 14-16.\\

\textbf{Rule 6: Safe speed}

\begin{quote}
    \textit{Every vessel shall at all times proceed at a safe speed so that she can take proper and effective action to avoid collision and be stopped within a distance appropriate to the prevailing circumstances and conditions.}
\end{quote}

\textbf{Rule 8: Action to avoid collision}

\begin{quote}
    \textit{(b) Any alteration of course and/or speed to avoid collision shall, if the circumstances of the case admit, be large enough to be readily apparent to another vessel observing visually or by radar; a succession of small alterations of course and/or speed should be avoided.}
    
    \textit{(c) If there is sufficient sea-room, alteration of course alone may be the most effective action to avoid a close-quarters situation provided that it is made in good time, is substantial and does not result in another close-quarters situation.}
    
    \textit{(d) Action taken to avoid collision with another vessel shall be such as to result in passing at a safe distance. The effectiveness of the action shall be carefully checked until the other vessel is finally past and clear.}
    
   \textit{ (e) If necessary to avoid collision or allow more time to assess the situation, a vessel shall slacken her speed or take all way off by stopping or reversing her means of propulsion.}
\end{quote}

\textbf{Rule 14: Head-on situation}

\begin{quote}
    \textit{(a) When two power-driven vessels are meeting on reciprocal or nearly reciprocal courses so as to involve risk of collision each shall alter her course to starboard so that each shall pass on the port side of the other.}
    
    \textit{(b) Such a situation shall be deemed to exist when a vessel sees the other ahead or nearly ahead and by night she could see the masthead lights of the other in a line or nearly in a line and/or both sidelights and by day she observes the corresponding aspect of the other vessel.}
    
    \textit{(c) When a vessel is in any doubt as to whether such a situation exists she shall assume that it does exist and act accordingly.}
\end{quote}

\textbf{Rule 15: Crossing situation}

\begin{quote}
     \textit{When two power-driven vessels are crossing so as to involve risk of collision, the vessel which has the other on her own starboard side shall keep out of the way and shall, if the circumstances of the case admit, avoid crossing ahead of the other vessel. }
\end{quote}
 
\textbf{Rule 16: Action by give-way vessel}

\begin{quote}
     \textit{Every vessel which is directed to keep out of the way of another vessel shall, so far as possible, take early and substantial action to keep well clear.}
\end{quote}

\subsection{Measures of collision risk}\label{sec:collision_risk}

The rules presented above are intended for human interpretation and contain ambiguities such as ``large enough'' (Rule 8) and ``substantial action'' (Rule 16). How can they be translated into a form suitable for reinforcement learning? An essential first step is recognizing the relationship between the COLREGs and collision risk. The COLREGs are in place to reduce collision risk and indirectly affect the risk level by influencing the probable behavior of the target ship (TS). Since there is a correlation between the rules and the risk level, employing a measure of risk as a proxy for the COLREGs may enable the RL agent to learn COLREG-compliant behavior. 

By analyzing the historical trends of measuring collision risk, three main developments can be observed (\cite{Xu2014}): traffic flow theory, ship safety domains, and collision risk indices. The initial efforts to quantify collision risk were based on \textit{traffic flow theory}, a method built on empirical studies and statistical traffic analysis in specific waters. For instance, \cite{cockcroft_1981} investigated the collision rates for ships of varying tonnage relative to their position in a water area. \cite{goodwin_1978} took it further and studied the rate of dangerous encounters. As statistical analysis of historical data was deemed insufficient for dynamic collision avoidance, \textit{ship safety domains} were introduced. The ship safety domain defines a region around the ship in question that other ships should not enter. Hence, there is a risk of collision if one ship is inside the safety domain of another, and the ship domain can be said to be a generalization of a safe distance (\cite{Szlapczynski2017}). When applying the ship domain to an encounter situation in order to determine risk, one of the four \textit{safety criteria} are normally used: 1) the OS domain should not be violated by a TS, 2) a TS domain should not be violated by the OS, 3) neither of the ship domains should be violated, or 4) ship domains should not overlap, such that they remain mutually exclusive. \cite{Rawson2014} and \cite{Wang2016} use the latter criterion of non-overlapping ship domains.

It is important to note that a ship domain is usually defined depending on which situation the ship finds itself in to respect the COLREGs. For instance, the domain used while the OS is overtaking another ship is symmetrical, with its origin coinciding with the center of the OS. Conversely, the origin is shifted to the right in a head-on situation, as close encounters on the starboard side should be avoided.

\begin{figure}[!ht]
\centering
  \includegraphics[width=0.8\linewidth]{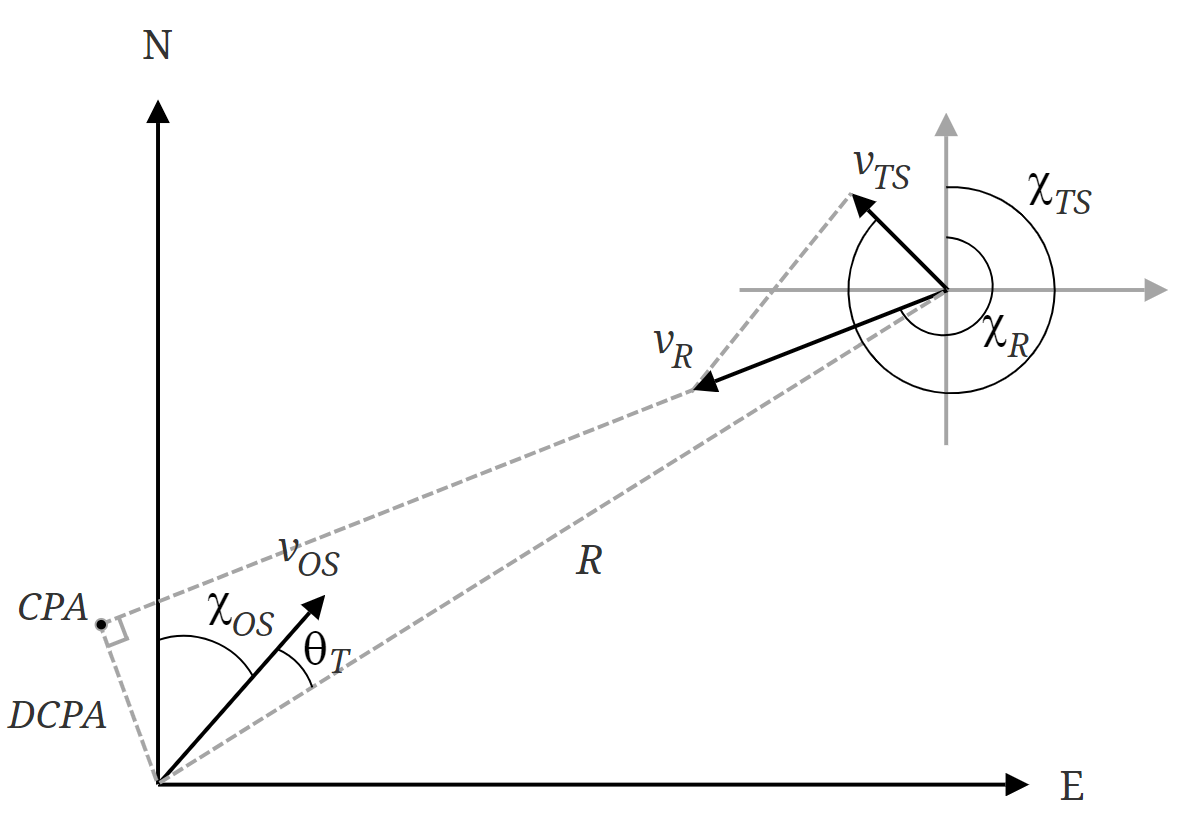}
  \caption[Geometrical representation of CPA and DCPA]{Geometric representation of CPA and DCPA.}
  \label{fig:DCPA_TCPA}
\end{figure}

\cite{davis_dove_stockel_1980}  expanded the theory of ship safety domains in their well-known work on ship arenas. The ship arena defines the distances around the OS at which action should be made to avoid a dangerous encounter and is, therefore, larger than the ship safety domains proposed initially. In addition to the OS's length and velocity, the distance to the closest point of approach (DCPA) and the time to the closest point of approach (TCPA) are used to construct the limits of the ship arena. A geometrical representation of DCPA and TCPA are presented in \Cref{fig:DCPA_TCPA}, giving rise to the equations

\begin{equation}\label{eq:DCPA}
   DCPA = R \sin(\chi_R - \chi_{OS} - \theta_T - \pi)  
\end{equation} 

and

\begin{equation}\label{eq:TCPA}
   TCPA = \frac{R}{V_R} \cos(\chi_R - \chi_{OS} - \theta_T - \pi)  
\end{equation} 

where $R$ is the absolute distance between the OS and TS, and $V_R$ and $\chi_R$ are the relative speed and course between them. In addition, $\chi_{OS}$ is the course of the OS, while $\theta_T$ is the bearing of the TS relative to the OS.

This leads to the subsequent development in collision risk evaluation, namely \textit{collision risk indices} (CRIs), which are primarily based on the DCPA and TCPA. In addition, a CRI can include the absolute distance from the OS to the TS $R$, velocity ratio $K$ of two encountering ships, relative course $\chi_R$, and other key features. Recently, simple CRIs alone are considered unable to capture collision risk's gradual and complex nature. As a result, combining the CRI with fuzzy logic or the fuzzy comprehensive evaluation method has become the norm. In fuzzy logic, fuzzy IF-THEN rules are applied to the parameters involved, such as DCPA and TCPA, to determine the risk level. In the fuzzy comprehensive evaluation method, on the other hand, \textit{membership functions}, $u(\cdot) \in [0,1]$, are used instead of IF-THEN rules, taking more details into account. The final CRI is then given as the weighted sum of the membership function outputs, as exemplified below:

\begin{subequations}\label{eq:ex_fuzzy}
\begin{align}
   CRI &= \alpha_{DCPA} \cdot u_{DCPA}(DCPA) \\
       &+  \alpha_{TCPA} \cdot u_{TCPA}(TCPA) \notag\\
       &+ \alpha_R \cdot u_R(R) \notag 
\end{align} 
\begin{equation}
  \alpha_{DCPA} + \alpha_{TCPA} + \alpha_R = 1
\end{equation}
\end{subequations}

\subsection{Deep reinforcement learning}

Model-free reinforcement learning (RL) methods train a decision-making agent through trial and error, where the agent is gathering experience from an environment supplying only a situational observation state and a corresponding reward. Applications of RL on high-dimensional, continuous control tasks heavily rely on function approximators to generalize over the state space. Even if classical, tabular solution methods such as Q-learning can be made to work (provided a discretizing of the continuous action space), this is not considered an efficient approach for control applications (\cite{lillicrap2015continuous}). In recent years, given their remarkable generalization ability over high-dimensional input spaces, the dominant approach has been the application of deep neural networks optimized using gradient methods. Several algorithms built on this principle have gained significant traction in the RL research community, most notably Deep Deterministic Policy Gradient (DDPG) (\cite{lillicrap2015continuous}), Asynchronous Advantage Actor Critic (A3C) (\cite{mnih2016asynchronous}), Proximal Policy Optimization (PPO) (\cite{schulman2017proximal}), and Soft Actor-Critic (SAC) (\cite{haarnoja2018SAC}). For continuous control tasks, this family of policy gradient methods is commonly considered the more efficient approach (\cite{tai2016survey}). Based on previous work, where the PPO algorithm significantly outperformed other methods on a learning problem similar to the one covered in this work (\cite{meyer_ASV_IEEE, larsen2021drlComparison}), we focus our efforts on this method.

\section{Methodology}
\label{section:methodology}
\subsection{Training environment}

\begin{figure}[ht!]
	\hspace*{-1cm}
	\centering
	\includegraphics[width=0.8\linewidth]{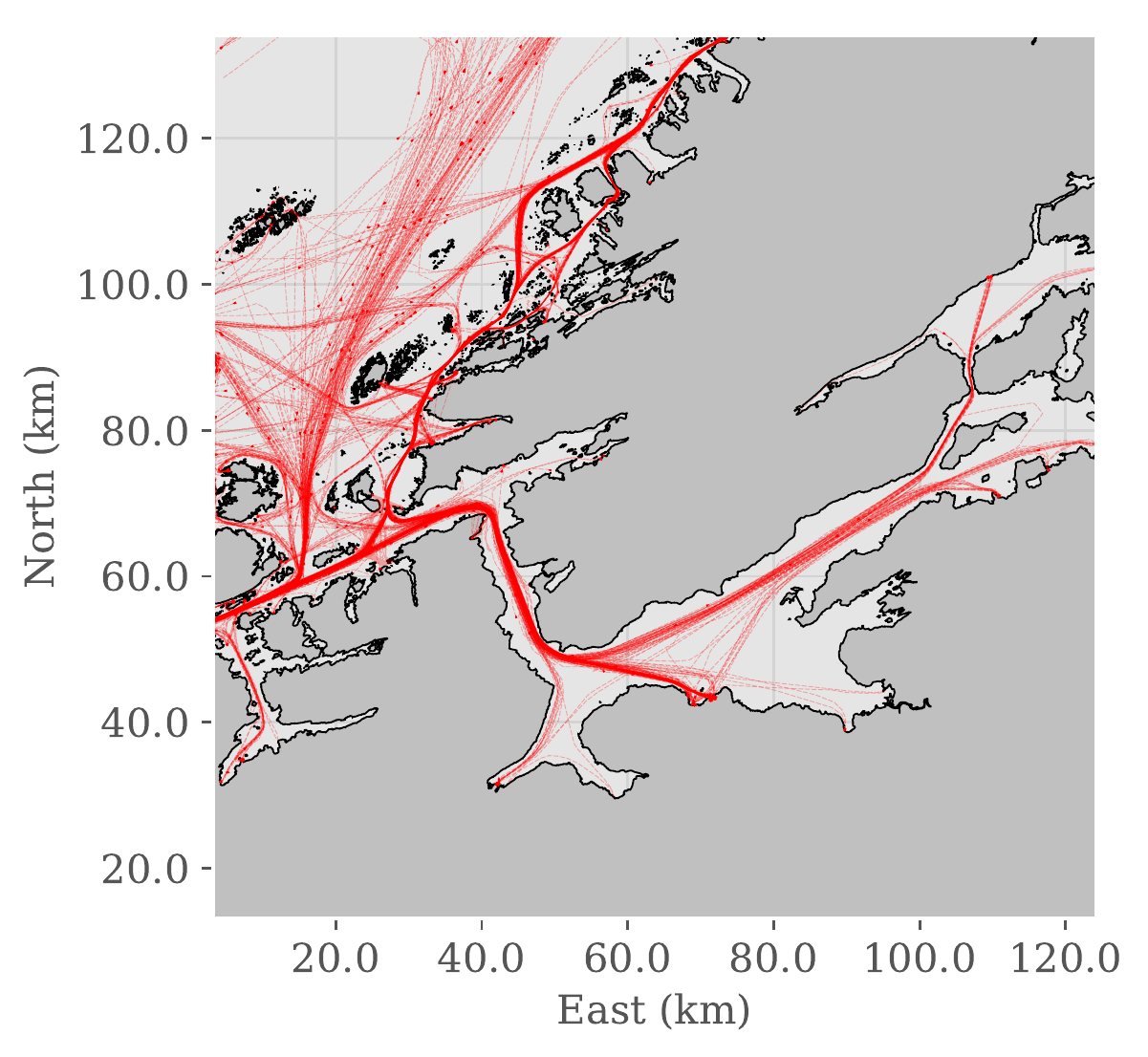}
	\caption[AIS data]{Snapshot of the marine traffic from 01.01.2020 to 06.02.2020 in the Trondheim fjord, based on AIS data. Each red line represents a recorded travel.}
	\label{fig:marinetraffic}
\end{figure}

\begin{figure}[ht!]
	\hspace*{-1cm}
	\centering
	\includegraphics[trim={0cm 0cm 0cm 0.5cm},clip,width=0.8\linewidth]{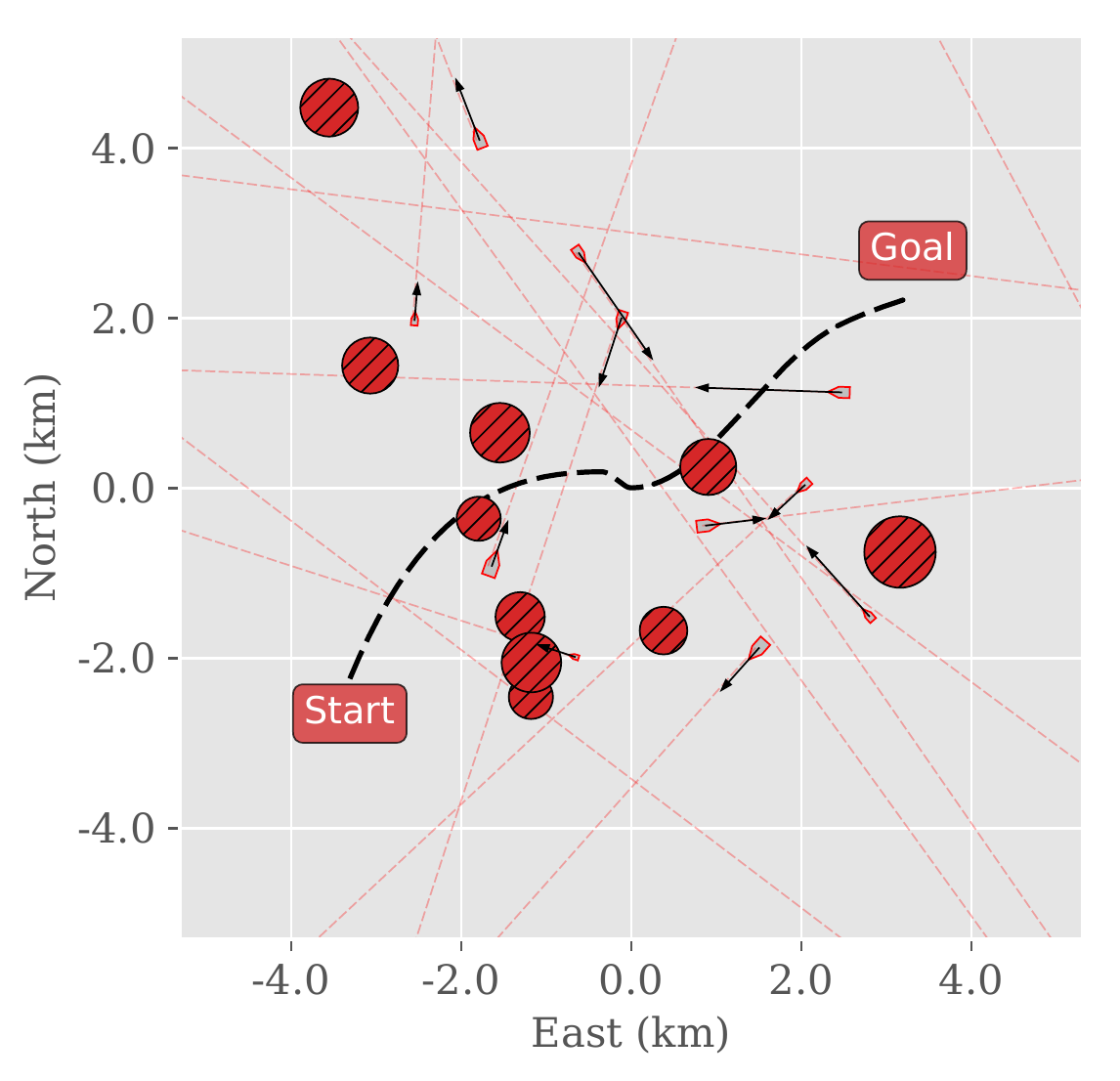}
	\caption[Random moving obstacles training scenario]{Random sample of the stochastically generated path following training scenario with moving obstacles. The circles are static obstacles representing landmasses, and the vessel-shaped objects are moving according to the trajectory lines and velocity vectors.}
	\label{fig:movingobstacles_demo}
\end{figure}

DRL-based autonomous agents have a remarkable ability to generalize their policy over the observation space, including the domain of unseen observations. Moreover, given the complexity and heterogeneity of the Trondheim Fjord environment, with archipelagos, shorelines, and skerries (see \Cref{fig:marinetraffic}), this ability will be fundamental to the agent's performance. However, the training environment in which the agent evolves from a blank slate to an intelligent vessel controller must be representative, challenging, and unpredictable to facilitate the generalization. If not for the generalization issues associated with this approach (\cite{codevilla2019exploring}), it would also allow the agent to train via behavior cloning based on historical AIS data. However, given the resolution of our terrain data, the resulting obstacle geometry is typically very complex, leading to overly high computational demands for simulating the functioning of the distance sensor suite. Moreover, the agent's perceptive observation space (\Cref{sec:methods:perception}) undergoes significant dimensionality reduction, resulting in the agent not benefiting from such high-frequency details in the simulation. Thus, the better choice is to craft an artificial training scenario with simple obstacle geometries. To reflect the dynamics of a real-world marine environment, we let the stochastic initialization method of the training scenario spawn other target vessels with deterministic, linear trajectories. Additionally, circular obstacles scattered around the environment substitute the real-world terrain. \Cref{fig:movingobstacles_demo} illustrates an instantiation of the training environment.

\subsection{Observation vector}
To facilitate the learning of a decision-making policy, the RL agent requires an observation vector, $s$, containing sufficient information about the vessel's state relative to the path in addition to situational sensor information. The complete observation vector is then constructed by concatenating navigation-based and perception-based features, which formally translates to $s = [s_{n}, s_{p}]^T$. In the context of this paper, we consider the term \textit{navigation} as the characterization of the vessel's state, i.e., its position, orientation, and velocity, with respect to the desired path. On the other hand, \textit{perception} refers to the observations made via the rangefinder sensor measurements. In the following, the path navigation feature vector, $s_{n}$, and the perceptive feature vector, $s_{p}$, are covered in detail.

\subsubsection{Navigation features}

A sufficiently information-rich path navigation feature vector would be such that it, on its own, could facilitate a satisfactory path-following controller. A few concepts often used in vessel guidance and control are helpful to formalize this. 
\begin{figure}[ht!]
\centering
    \includegraphics[width=0.8\linewidth]{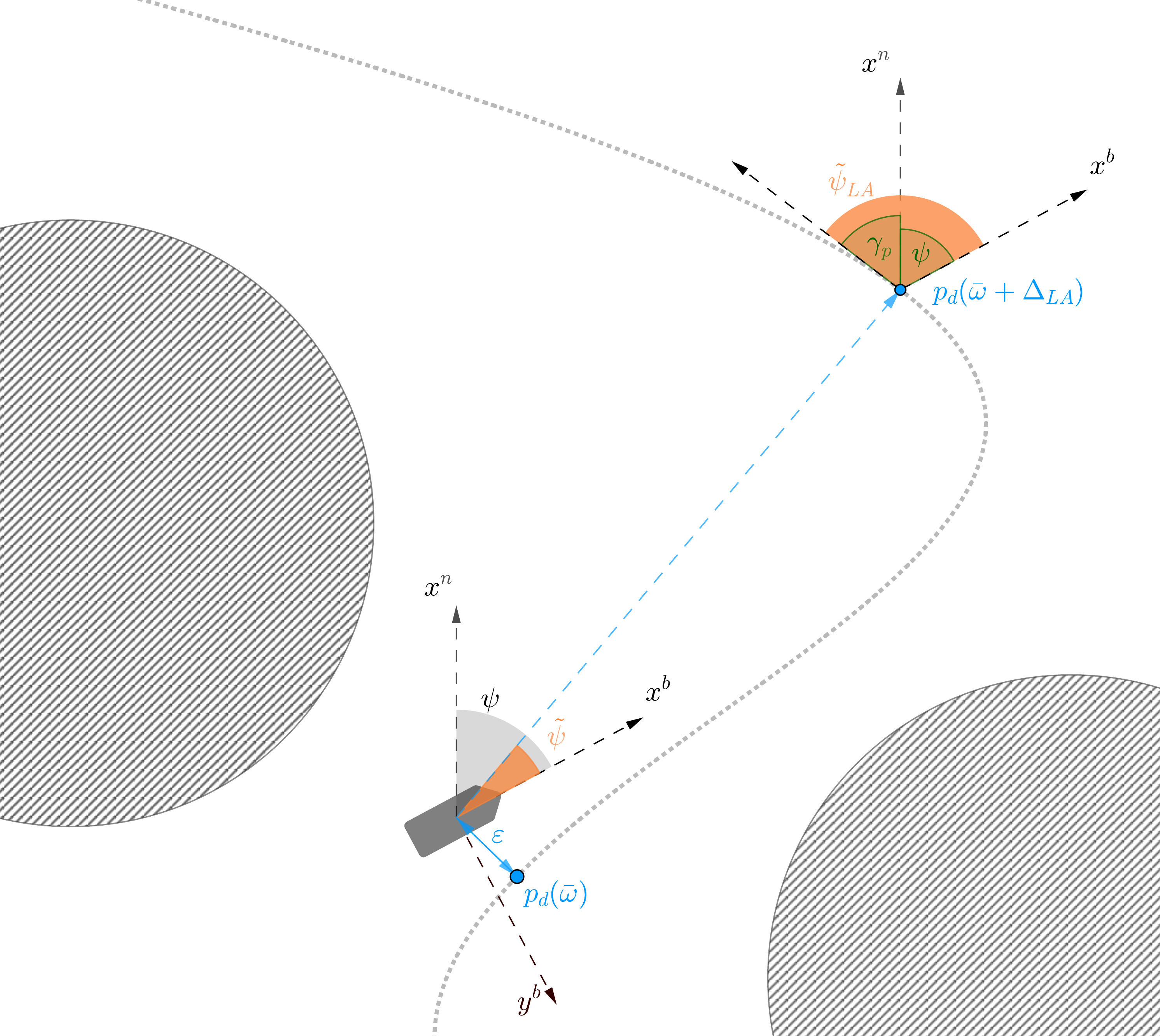}
	\caption[Path following navigation]{Illustration of key path-following concepts in vessel guidance and control. The path reference point, $\bm{p}_d(\bar{\omega})$, describes the point on the path with the closest Euclidean distance to the vessel, while the look-ahead reference point, $\bm{p}_d (\bar{\omega} + \Delta_{LA})$, is located a distance, $\Delta_{LA}$, further along the path.}
	\label{fig:navigationfig}
\end{figure}
First, we introduce the mathematical representation of the parameterized path, which is expressed as
\begin{equation}
\bm{p}_d(\omega) = \left[x_d(\omega), y_d(\omega)\right]^T
\end{equation}
\noindent where $x_d(\omega)$ and $y_d(\omega)$ are defined in the NED frame. Navigating the path necessitates a reference point, which is continuously updated based on the vessel's position. We define this reference point as the point on the path that has the closest Euclidean distance to the vessel, given its current position, as illustrated in \Cref{fig:navigationfig}. To find this, we calculate the corresponding value of the path variable $\bar{\omega}$ at each time step. This is an equivalent problem formulation because the path is defined implicitly by the value of $\omega$. Formally, this translates to the optimization problem
\begin{equation}
\bar{\omega} =  \argmin_\omega \left(x^n - x_d(\omega)\right)^2 + \left(y^n - y_d(\omega)\right)^2,
\end{equation}
which, using the Newton–Raphson method, can be calculated accurately and efficiently at each time step. We define the corresponding Euclidean distance to the path, i.e., the deviation between the desired path and the current track, as the cross-track error (CTE) $\epsilon$. Formally, we thus have that
\begin{equation}
\epsilon =  \left\lVert \left[ x^n, y^n \right]^T - \bm{p}_d(\bar{\omega}) \right\rVert.
\end{equation}
Next, we consider the look-ahead point, $\bm{p}_d (\bar{\omega} + \Delta_{LA})$, to be the point that lies a constant distance further along the path from the reference point $\bm{p}_d(\bar{\omega})$. Look-ahead based steering, i.e., setting the look-ahead point direction as the desired course angle, is a commonly used guidance principle (\cite{fossen11}). The look-ahead distance, $\Delta_{LA}$, is set by the user and controls how aggressively the vessel should reduce the distance to the path. 

We then define the heading error, $\tilde{\psi}$, as the change in heading needed for the vessel to navigate straight towards the look-ahead point from its position, as illustrated in \Cref{fig:navigationfig}. Formally, $\tilde{\psi}$ is defined as
\begin{equation}
\tilde{\psi} = \atantwo{\left(\frac{y_d(\bar{\omega} + \Delta_{LA}) - y^n}{x_d(\bar{\omega} + \Delta_{LA}) - x^n}\right)} - \psi,
\end{equation}
where $\psi$ is the vessel's heading and $x^n, y^n$ are the NED-frame vessel coordinates as defined earlier.

However, even if minimizing the heading error will yield good path adherence, taking into account the path direction at the look-ahead point might improve the smoothness of the resulting vessel trajectory. Referring to the first-order path derivatives as $x_{p}^{\prime}(\bar{\omega})$ and $y_{p}^{\prime}(\bar{\omega})$, we have that the path angle, $\gamma_p$, in general, can be expressed as a function of arc-length, $\omega$, such that
\begin{equation}
\gamma_{p}(\bar{\omega}) = \atantwo{(y_{p}^{\prime}(\bar{\omega}), x_d^{\prime}(\bar{\omega}))}.
\end{equation}
As visualized in \Cref{fig:navigationfig}, the path direction at the look-ahead point is then given by $\gamma_{p}(\bar{\omega} + \Delta_{LA})$. We then define the look-ahead heading error, which is zero in the case when the vessel is heading in a direction that is parallel to the path direction at the look-ahead point, as
\begin{equation}
\tilde{\psi}_{LA} = \gamma_{p}(\bar{\omega} + \Delta_{LA}) - \psi
\end{equation}
Our assumption is then that the navigation feature vector $s_{n}$, defined as outlined in \Cref{tab:obs_vector_path}, should provide a sufficient basis for the agent to intelligently adhere to the desired path. The navigation features are then formally defined as
\begin{equation}
s_{n}^{(t)} = \left[u^{(t)}, v^{(t)}, r^{(t)}, \epsilon^{(t)}, \tilde{\psi}^{(t)}, \tilde{\psi}_{LA}^{(t)}\right]^T.
\end{equation}
\begin{table}[ht!]
	\scriptsize
	\centering
	\caption[Path-following feature vector]{Path-following feature vector $s_{n}$ at timestep $t$.}\label{tab:obs_vector_path}
	\begin{tabular}{ll}
		\hline
		\textbf{Feature} & \textbf{Definition}\\
		\hline
		Surge velocity & $u^{(t)}$ \\
		Sway velocity & $v^{(t)}$ \\
		Yaw rate & $r^{(t)}$ \\
		Cross-track error & $\epsilon^{(t)}$ \\
		Heading error & $\tilde{\psi}^{(t)}$ \\
		Look-ahead heading error & $\tilde{\psi}_{LA}^{(t)}$ \\
		\hline
	\end{tabular}
\end{table}

\subsubsection{Perception features} \label{sec:methods:perception}
Using a set of rangefinder sensors as the basis for obstacle avoidance is a natural choice, as it yields a comprehensive yet intuitive representation of any neighboring obstacles. This configuration should also enable a relatively straightforward transition from the simulated environment to a real-world one, given that rangefinder sensors such as lidars, radars, sonars, or depth cameras are commonly used

In our setup, the vessel is equipped with $N$ distance sensors with a maximum detection range of $S_r$, distributed uniformly with $360\degree$ coverage. While the area behind the vessel is obviously of lesser importance, e.g., unnecessary to consider when navigating purely static terrain, the possibility of overtaking situations where the agent must react to another vessel approaching from behind makes full sensor coverage necessary. The most natural approach to constructing the final observation vector would be to concatenate the path information feature vector with the array of sensor outputs. However, initial experiments with this approach resulted in the training process stagnating at an unsatisfactory agent performance level. A likely explanation for this failure is the size of the observation vector, which was fed to the agent's fully connected policy and value networks; as the input size becomes large, the agent suffers from the well-known \textit{curse of dimensionality}. Due to the resulting network complexity and the exponential relationship between the dimensionality and volume of the observation space, the agent fails to generalize new, unseen observations intelligently (\cite{goodfellowDL}). An obvious solution is to reduce the observation space's dimensionality significantly. However, simply reducing the resolution is infeasible, as this would accordingly degrade the agent's situational awareness.

In this work, we partition the sensor suite into $D$ sectors, each of which produces a scalar measurement included in the final observation vector, effectively summarizing the local sensor readings within the sector. However, given our desire to minimize its dimensionality, dividing the sensors into sectors of uniform size is sub-optimal as obstacles located in front of the vessel are significantly more critical and thus require higher perceptive accuracy than those located at its rear. In order to realize such a non-uniform partitioning, we use a logistic function - a choice that also fulfills our general preference for symmetry. Assuming a counter-clockwise ordering of sensors and sectors starting at the rear of the vessel, we map a given sensor index, $i \in \integerset{N}$, to a sector index, $k \in \integerset{D}$, according to
\begin{equation}
\kappa : i \mapsto \kappa(i) = \floor*{\underbrace{D \sigma\left(\frac{\gamma_C i}{N} - \frac{\gamma_C}{2} \right)}_\text{Non-linear mapping} - \underbrace{D \sigma \left( - \frac{\gamma_C}{2} \right)}_\text{Constant offset}},
\end{equation}
where $\sigma$ is the logistic sigmoid function, and $\gamma_C$ is a scaling parameter controlling the density of the sector distribution such that decreasing it will yield a more evenly distributed partitioning. We can then formally define the distance measurement vector for the $k^{th}$ sector, which we denote by $\bm{w}_k$, according to
\begin{align*}
\bm{w}_{k, i} &=x_i&     & \text{ for } i \in \integerset{N} \text{ such that } \kappa(i) = k
\end{align*}
\begin{figure}[ht!]
	\hspace*{0.4cm}
	\centering
	\includegraphics[trim={0cm 0cm 0.0cm 0.0cm},clip,width=0.9\linewidth]{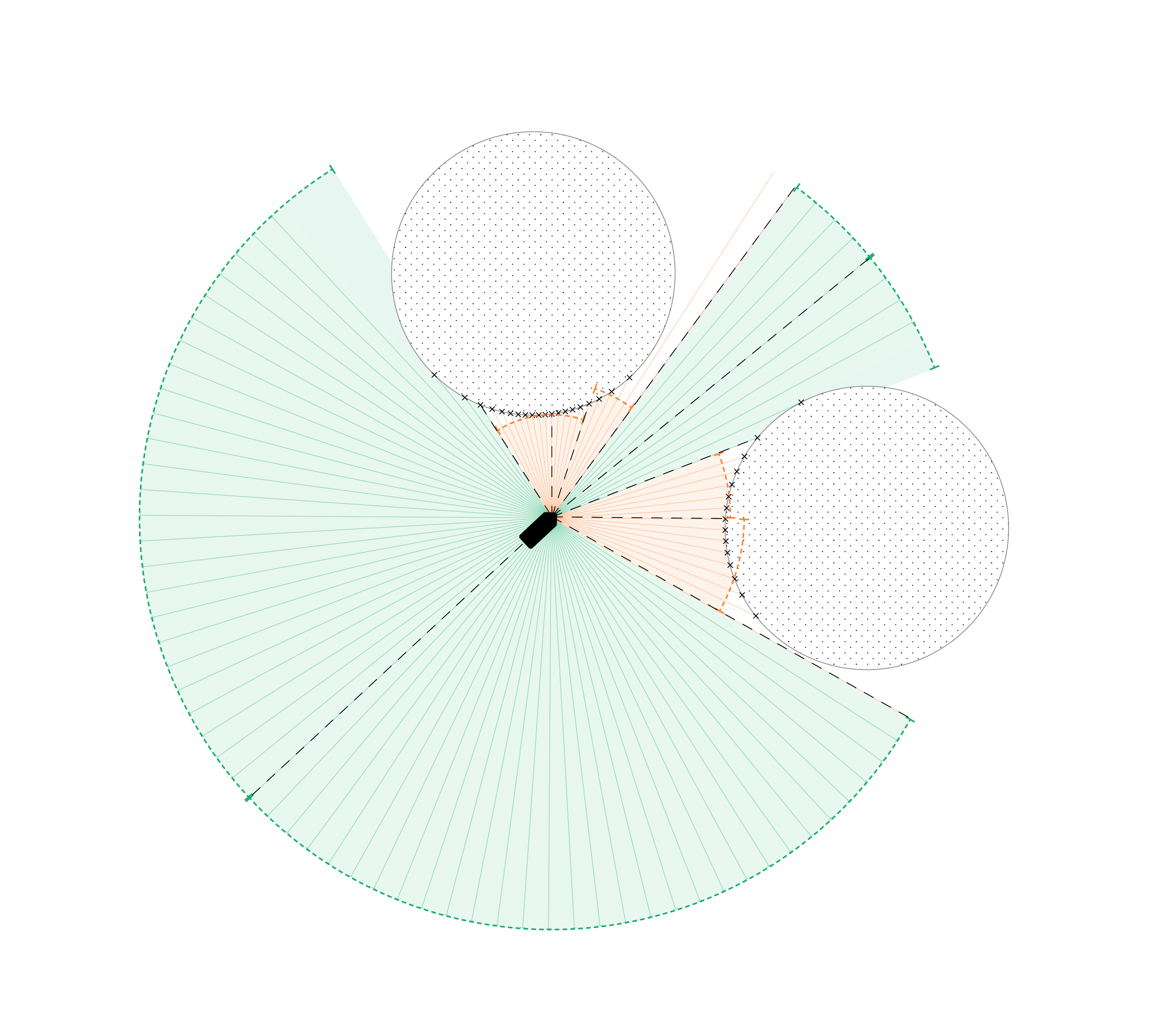}
	\caption[Sector-partitioned rangefinder sensor suite]{Rangefinder sensor suite containing $N$ distance sensors, partitioned into $D$ sectors (black dashed lines) according to the mapping function $\kappa$. The dashed edges illustrate the maximum reachable distance in each sector, as calculated by the feasibility pooling algorithm. The perceptive distance component of the RL-agent's observation space consists of the closeness mapping of these distances.}
	\label{fig:sensorswithsectors}
\end{figure} 

Next, we select a mapping $f : \mathbb{R}^n \mapsto \mathbb{R}$, which takes the vector of distance measurements $\bm{w}_k$, for an arbitrary sector index $k$, as input, and outputs a scalar value based on the current sensor readings within that sector. The \textit{feasibility pooling} procedure, introduced in \cite{meyer_ASV_IEEE}, calculates the maximum reachable distance within each sector, taking into account the obstacle sensor readings' location and the vessel's width. This method iterates over the sector's distance measurements in ascending order and checks whether it is feasible for the vessel to advance beyond this level. As soon as the broadest available opening within a distance level is deemed too narrow given the vessel's width, the maximum reachable distance has been reached. Formally, we define $f$ as the feasibility pooling algorithm, and the resulting perceptive distance observation is summarized in \Cref{fig:sensorswithsectors}.
To finalize the processing of distance measurements, we introduce the concept of \textit{closeness}. An obstacle's closeness is zero if it is at a distance further than $S_r$ away from the vessel and unity if the vessel has collided with the obstacle. Furthermore, within this range, is it reasonable to map distance to closeness in a logarithmic fashion, such that, following human intuition, the difference between $10$m and $100$m is more significant than the difference between, for instance, $510$m and $600$m. Formally, the maximum reachable distance, $d$, maps to closeness, $c(d) : \mathbb{R} \mapsto [0, 1]$, according to
\begin{equation} \label{eq:closeness}
c(d) = \text{clip}\left(1 - \frac{\log{(d+1)}}{\log{(S_r+1)}}, \;0, \;1\right).
\end{equation}

\subsubsection{Motion detection}
The maximum reachable distance in a sector may equal the maximum sensor range even though there is an obstacle in that sector. Thus, by applying the feasibility pooling algorithm to reduce the dimensionality of the rangefinder suite, the resulting closeness observation may fail to inform the RL agent about nearby obstacles. To make the agent aware of nearby moving obstacles, we incorporate the velocities of the nearest obstacle in each sector into the observation vector. Admittedly, while this implementation is trivial in a simulated environment, a real-world implementation will necessitate a reliable way of estimating obstacle velocities based on sensor data. However, even if this can be challenging due to uncertainty in the sensor readings, object tracking is a well-researched computer vision discipline. We reserve the implementation of such a method to future research but refer the reader to \cite{granstrom2016extendedobjecttracking} for a comprehensive overview of the current state of this field.

Specifically, the decomposition, which yields the x and y component of the obstacle velocity, considers the coordinate frame in which the y-axis is parallel to the centerline of the sensor sector in which the obstacle is present. Thus, we provide the decomposed velocity of the closest moving obstacle within each sector as features for the agent's observation vector. For each sector $k$, we denote the corresponding decomposed $x$ and $y$ velocities as $v_{x, k}$ and $v_{y, k}$, respectively. Naturally, if no moving obstacles are present within the sector, both components are zero.

\subsubsection{Perception state vector}
By concatenating the closeness of the maximum reachable distance and the decomposed obstacle velocity for each sector, we then define the perception state vector, $s_{p}$, as
\begin{equation}
s_{p}^{(t)} = \left[ \underbrace{c\left(\left(\bm{w}_1^{(t)}\right)\right),\; v_{x, 1}^{(t)},\; v_{y, 1}^{(t)}}_\text{First sector},\; \dots\right]^T.
\end{equation}

\subsection{Risk-based implementation of COLREGs}\label{sec:Quantitative_implementation}
In model-free RL, the trained agent will assume a policy that maximizes the expected reward. To lead this policy to adhere to the COLREGs, we must incorporate them into the reward function. As previously mentioned, the rules are ambiguous and cannot be implemented explicitly. Instead, we use collision risk indices (CRIs) as analogs, and the following motivates how they are intended to guide the RL agent towards COLREG-compliance. 

\subsubsection{Risk-based reward function}

Building on the theory presented in \Cref{sec:collision_risk}, a collision risk index (CRI) is calculated using fuzzy evaluation. Here, this translates to a weighted sum of evaluated risk factors, a method described in detail in \Cref{sec:calc_col_risk}. This method encapsulates collision risk's continuous and fuzzy nature, making it a convincing choice for translating the COLREGs into a DRL-based framework. Collision risk is typically only applied to encounter situations between two dynamic objects, and the collision risk index to be presented here is no exception. Thus, the reward components for path following, static obstacle avoidance, collision penalty, and living penalty must be defined separately. The corresponding components from a previous approach (\cite{Meyer2020}) are applied here due to the excellent path following and obstacle avoidance results. The reward components for path following and static obstacle avoidance are given in \Cref{eq:path_following,eq:r_coll_stat}, while the collision and living penalties are negative constants. As a result, the total reward function has the same structure, reiterated in \Cref{eq:preliminary_reward_function_2}, except for a risk-based penalty for dynamic obstacles ($r_{colav,dyn}$). 

\begin{equation}\label{eq:path_following}
r_{path}^{(t)} =  \underbrace{\left(\tfrac{u^{(t)}}{U_{max}} \cos{\tilde{\psi}^{(t)}} + \gamma_r \right)}_\text{Velocity-based reward} \underbrace{\left(\exp{\left(-\gamma_{\epsilon} |\epsilon^{(t)}|\right)} + \gamma_r\right)}_\text{CTE-based reward} -\gamma_r^2
\end{equation}

\begin{equation}\label{eq:r_coll_stat}
r_{colav, stat}^{(t)} = -\frac
{\textstyle \displaystyle\sum_{i=1}^{N}{\frac{\alpha_x}{1+\gamma_{\theta, stat} |\theta_i|}  \exp{(-\gamma_x x_i )} }}
{\textstyle \displaystyle\sum_{i=1}^{N}{\frac{1}{1+\gamma_{\theta, stat} |\theta_i|}}}
\end{equation}

\begin{equation}\label{eq:preliminary_reward_function_2}
r = 
\begin{cases}
    r_{collision}, & \text{if collision}\\
   \lambda r_{path} + \left( 1 - \lambda \right) r_{colav} + r_{exists}, & \text{otherwise}
\end{cases}
\end{equation}

The penalty for dynamic obstacles makes part of the overall penalty for collision avoidance, denoted $r_{colav}$ and given by 

\begin{equation}
    r_{colav} = r_{colav,dyn} + r_{colav,stat}.
\end{equation}\label{eq:r_colav_2}

For every TS within the OS's sensor range, a collision risk index (CRI) $\in [0,1]$ is calculated (see \Cref{sec:calc_col_risk}). Since the CRI increases proportionally to collision risk, it can be used semi-directly in the reward function. By multiplying the $CRI_i$ of each target vessel, $i$, with a scaling factor $\beta_{CRI} > 0$, the penalty level can be weighted relative to the rest of the reward function:

\begin{equation}
    r_{colav,dyn} = - \sum \beta_{CRI} \cdot CRI_i
\end{equation}\label{eq:preliminary_reward_function_3}

\subsubsection{Calculating the collision risk index}\label{sec:calc_col_risk}

In order to determine the collision risk in an encounter situation, one must first define what constitutes a collision risk and how much each risk factor contributes to the overall risk. The state-of-the-art methods of computing CRIs generally use fuzzy evaluation (\cite{Xu2014}), making it a natural choice here too. In short, three steps should be followed:\\

\begin{enumerate}
    \item Define individual risk factors.
    \item Define membership functions.
    \item Design overall CRI as a function of membership functions.\\
\end{enumerate}

The chosen risk factors and their membership functions are elaborated on in the following, leading up to the CRI function design.

A common starting point for defining risk is looking at the distance and time to the point of closest approach, denoted DCPA and TCPA. As the descriptive name suggests, the closest point of approach (CPA) is the closest point relative to the OS that the TS in question will come, given that the relative course and relative velocity between the two ships stay the same. The DCPA, then, is the distance to the CPA, while the TCPA is the time until the TS arrives at the CPA. Put differently, the DCPA quantifies the severity of a potential collision situation, while the TCPA quantifies its urgency. When determining the risk level associated with them, it is customary to employ upper and lower bounds for these quantities, denoted $d_L$ and $d_U$ for DCPA, and $t_L$ and $t_U$ for TCPA. Doing so, the membership functions $u_{DCPA}$ and $u_{TCPA}$ output unity (highest risk level) whenever $|\text{DCPA}| \leq d_L$ and $|\text{TCPA}| \leq t_L$, respectively. Conversely, their outputs are zero when $|\text{DCPA}| \geq d_U$ and $|\text{TCPA}| \geq t_U$. As was done in \cite{Gang2016a}, a second-order function is used between the two extremities. \cite{Chen2014} use a sinusoidal function instead. Although the latter has the virtue of being smooth, it was deemed inexpedient due to the large outputs for a wide interval of values, overshadowing other elements of the CRI. Since the sensor range used in this work is relatively short (1500 m), the steeper second-order function improved learning. It is worth noting that the sinusoidal function may be more suited in a setup with fewer obstacles and vessels where AIS data from a larger region is used. 

\begin{figure}[ht!]
\centering
	\hspace*{0cm}\includegraphics[trim={0cm 0cm 0.0cm 0.0cm},clip,width=0.8\linewidth]{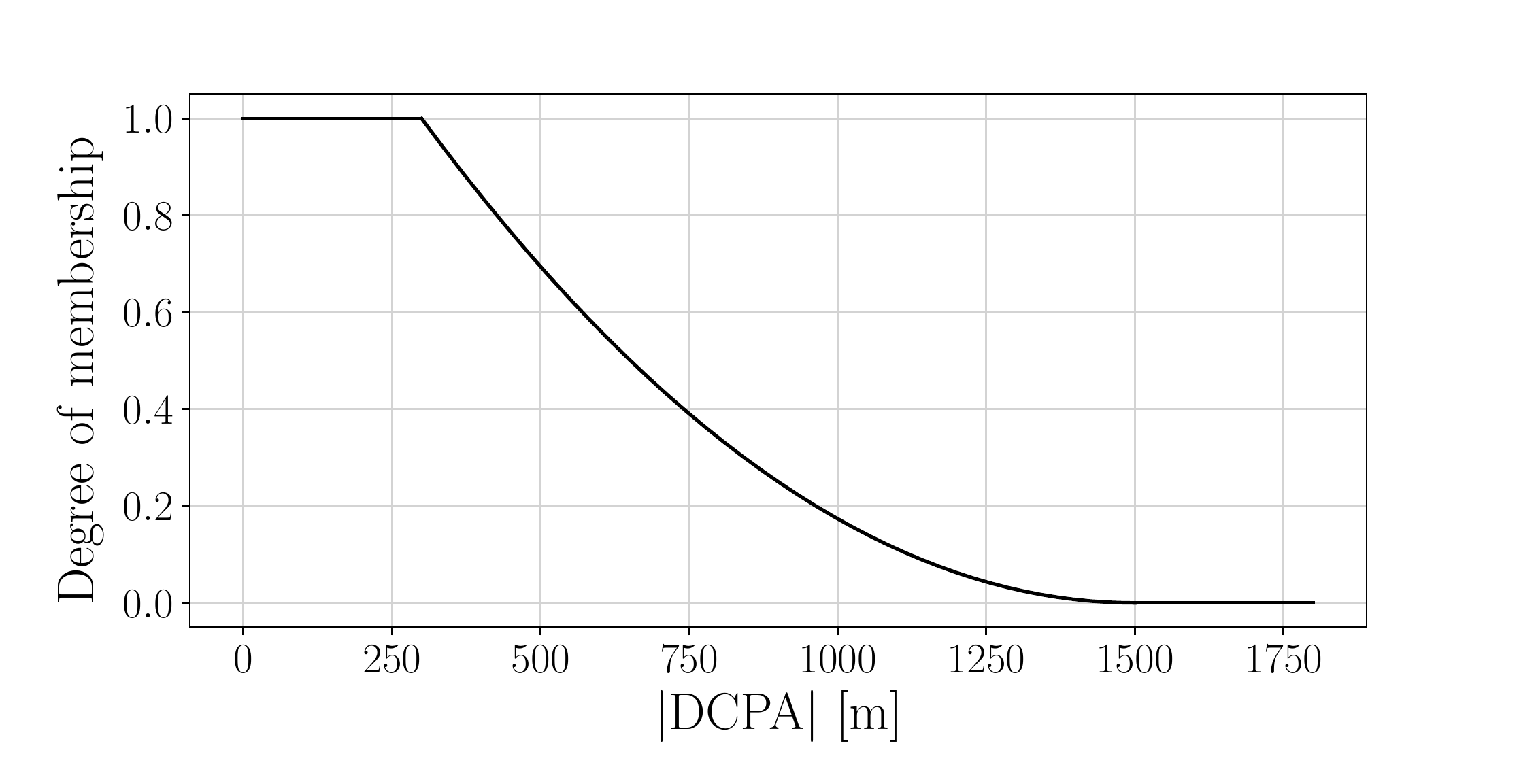}
	\caption[DCPA membership function]{Membership function for DCPA with $d_L$ = 320 m and $d_U$ = 1500 m.}
	\label{fig:mem_DCPA}
\end{figure}

The values for the lower and upper bounds depend largely on the application. In general, $d_L$ defines the minimal safe encounter distance, and $d_U$ is the absolute safe encounter distance (\cite{Gang2016a}). For DCPA, the membership function is defined as

\begin{equation}
u_{DCPA} = 
\begin{cases}
    1 & \text{if $|DCPA| \leq d_L$}\\
    0 & \text{if $|DCPA| \geq d_U$}\\
    \left(\frac{d_U - |DCPA|}{d_U - d_L}\right)^2 & \text{otherwise}\\
\end{cases}
\end{equation}\label{eq:uDCPA}

with $d_L$ and $d_U$ as positive integers. The DCPA membership function is presented graphically in \Cref{fig:mem_DCPA}.

For the bounds on TCPA, the method used in \cite{Gang2016a} and presented in \Cref{eq:TCPA_bounds} is employed. Doing so adjusts the output of $u_{TCPA}$ according to the distance between the OS and TS, accurately presenting the high risk when the distance is below or close to the lower bound $d_L$ and low risk when it is closer to the upper bound $d_U$. It is assumed that DCPA never exceeds $d_U$, meaning that $d_U$ is set to the maximum detectable DCPA.

\begin{subequations}\label{eq:TCPA_bounds}
\begin{equation}
    t_L = 
    \begin{cases}
        \frac{\sqrt{d_L^2 - DCPA^2}}{v_R} & \text{if $|DCPA| \leq d_L$}\\
        \frac{d_L - DCPA}{v_R} & \text{if $|DCPA| > d_L$}\\
    \end{cases}
\end{equation}
\begin{equation}
    t_U = \frac{\sqrt{d_U^2 - DCPA^2}}{v_R} 
\end{equation} 
\end{subequations}

In \cite{Gang2016a}, equal importance has been given to positive and negative values of TCPA through the membership function below:

\begin{equation}
u_{TCPA} = 
\begin{cases}
    1 & \text{if $|TCPA| \leq t_L$}\\
    0 & \text{if $|TCPA| \geq t_U$}\\
    \left(\frac{t_U - |TCPA|}{t_U - t_L}\right)^2 & \text{else}\\
\end{cases}
\end{equation}\label{eq:uTCPA_Gang}

\begin{figure}[ht!]
\centering
	\hspace*{0cm}\includegraphics[trim={0cm 0cm 0.0cm 0.0cm},clip,width=0.7\linewidth]{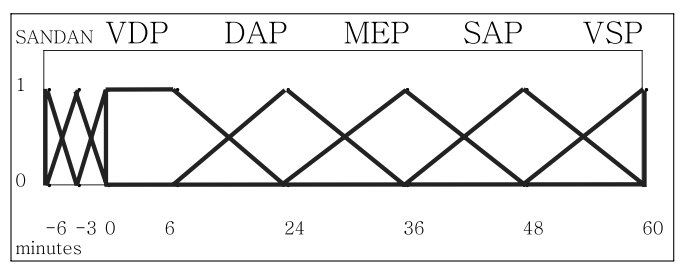}
	\caption[Example fuzzy membership function for TCPA]{Membership function for TCPA employed in \cite{Park2006}. SAN = Safe Negative, DAN = Dangerous Negative, VDP = Very Dangerous Positive, DAP = Dangerous Positive, MEP = Medium Positive, SAP = Safe Positive, VSP = Very Safe Positive.}
	\label{fig:fuzzy_TCPA}
\end{figure}

However, noting that negative values of TCPA indicate that the OS and TS have passed each other, it makes sense to pay attention to the sign of TCPA. This is supported by \cite{Park2006}, where a fuzzy case-based reasoning system for collision avoidance is proposed. In their work, the TCPA membership function in \Cref{fig:fuzzy_TCPA} is applied, indicating the significantly higher risk associated with positive values of TCPA.

\begin{figure}[ht!]
\centering
	\hspace*{0cm}\includegraphics[trim={0cm 0cm 0.0cm 0.0cm},clip,width=0.8\linewidth]{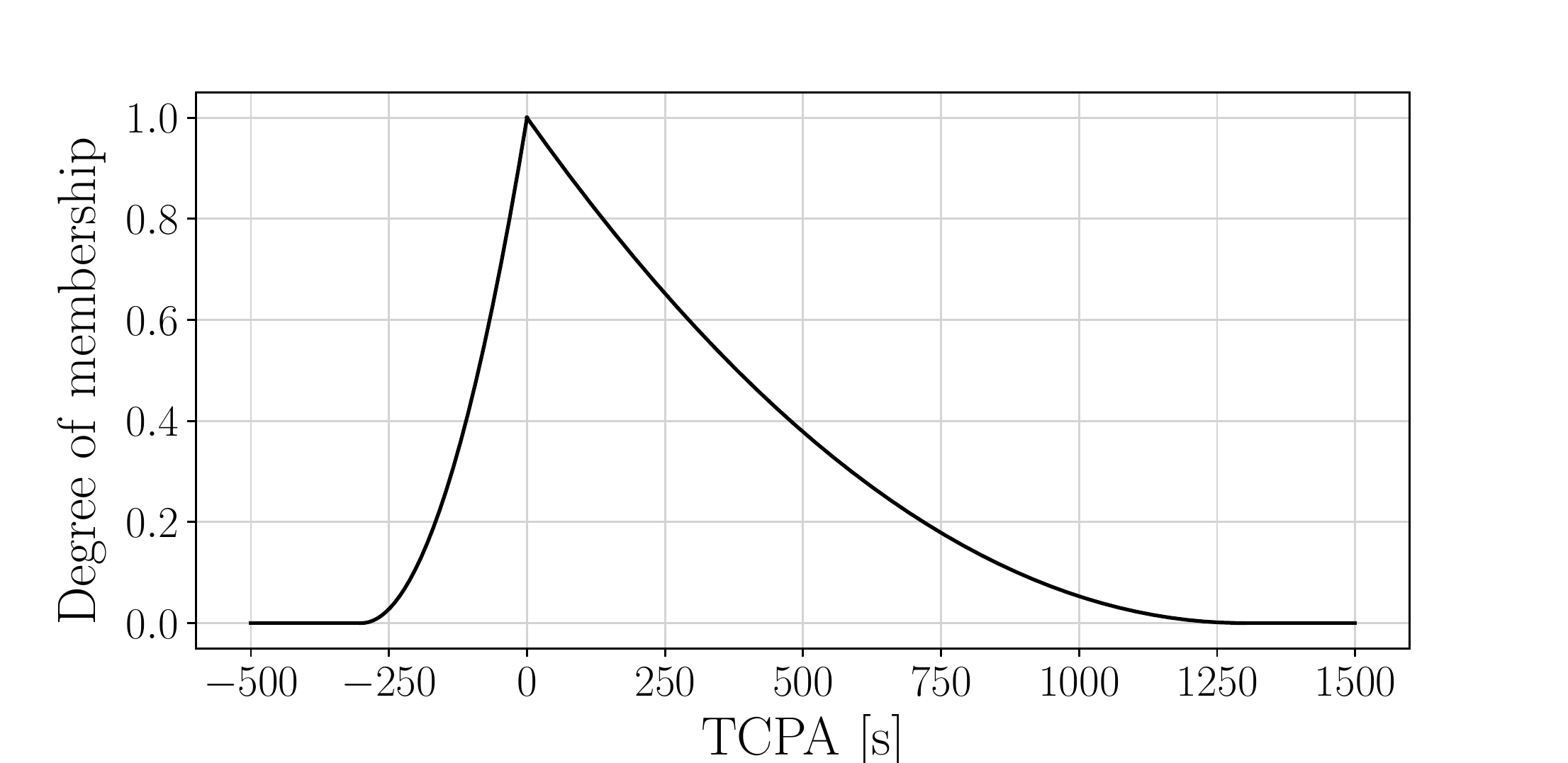}
	\caption[TCPA membership function]{Membership function for TCPA with $d_L$ = 320 m, $d_U$ = 1500 m and $v_R$ = 1 m/s.}
	\label{fig:mem_TCPA}
\end{figure}

Following this line of reasoning, a distinction between positive and negative values of TCPA is made according to \Cref{eq:u_TCPA}. The cut-off value for negative values (negative limit) was chosen as $t_{NL} = \frac{d_L}{v_R}$, such that the degree of membership is larger than zero whenever the OS is less than $t_{NL}$ time steps away from the TS. The membership function for TCPA is plotted in \Cref{fig:mem_TCPA}.

\begin{equation}\label{eq:u_TCPA}
  u_{TCPA}=
  \begin{cases}
    \begin{cases}
      1 & \text{if $TCPA \leq t_L$}\\
      0 & \text{if $TCPA \geq t_U$}\\
      \left(\frac{t_U - TCPA}{t_U - t_L}\right)^2 & \text{else}\\
    \end{cases}
    & \text{if $TCPA \geq$ 0}\\
    \begin{cases}
       0 & \text{if $TCPA \leq t_L$}\\
       \left(\frac{t_{NL} - |TCPA|}{t_{NL}}\right)^2 & \text{else}\\
    \end{cases}
    &\text{if $TCPA < 0$}\\
  \end{cases}
\end{equation}

%
%

Further, the collision risk depends on the position of the TS relative to the OS, which can be expressed through the absolute distance, $R$, between them and the bearing angle of the TS, $\theta_T$. Since the risk is higher on the starboard side of the OS, as expressed in Rule 14 (head-on situation) of the COLREGs, the membership functions should be designed with a bias on that side. Inspired by \cite{davis_dove_stockel_1980}, it is customary to introduce a bias of $19\degree$ starboard. Davis developed the concept of ship arena, briefly described in \Cref{sec:collision_risk}, and designed a scaling of the upper bound:

\begin{equation}\label{eq:R_D}
R_{D} = 
1.7\cos\left(\theta_T\frac{\pi}{180}-19\degree\right)+\sqrt{\left(4.4+2.89\cos^{2}\left(\theta_T\frac{\pi}{180}-19\degree\right)\right)},
\end{equation}

while the lower bound is usually 12 times the OS length $L_{pp}$ (\cite{Gang2016a}) but set to 8$L_{pp}$ here due to the smaller scale. Initially, the upper bound given by $R_D$ was implemented, but it quickly became apparent that adjustments had to be made to ensure that the agent received sufficiently negative reward when approaching TSs, regardless of their bearing angle. The difference in scaling of 4.4 times for ships detected at 19$\degree$ and 161$\degree$ (180$\degree - 19\degree$) was too large considering the relatively densely populated training and testing scenarios and a restricted sensor range of 1500 m. Through testing, it was observed that the distance membership function could be made uniform while still preserving the correct behavior in head-on situations as long as the membership function for the bearing angle, $\theta_T$, was given enough weight. As a result, the lower and upper bounds for the absolute distance, $R$, were chosen as

\begin{subequations}\label{eq:R_bounds}
\begin{equation}
   R_L =\beta_{RL} L_{pp}
\end{equation} 
\begin{equation}
  R_U = \beta_{RU} L_{pp}
\end{equation}
\end{subequations}

with $\beta_{RL}$ and $\beta_{RU}$ chosen as appropriate scaling constants.

\begin{figure}[ht!]
\centering
	\hspace*{0cm}\includegraphics[trim={0cm 0cm 0.0cm 0.0cm},clip,width=0.8\linewidth]{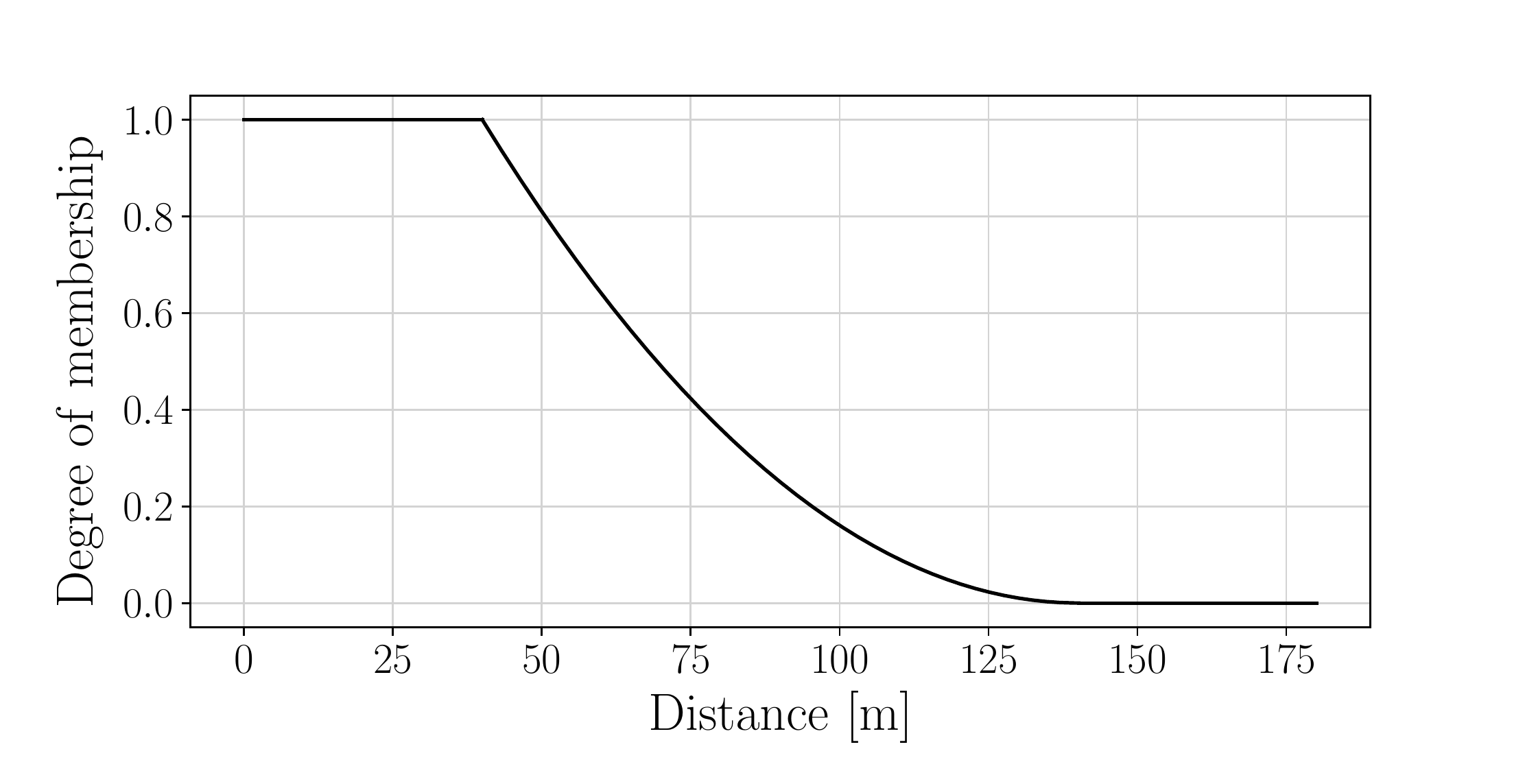}
	\caption[Distance membership function]{Membership function for distance to the target ship, with $\theta_T$ = 0$\degree$.}
	\label{fig:mem_dist}
\end{figure}

Following the logic applied to the membership functions for TCPA and DCPA, we arrive at the following membership function for the absolute distance between the OS and TS:

\begin{equation}\label{eq:uR}
u_{R} = 
\begin{cases}
    1 & \text{if $R \leq R_L$}\\
    0 & \text{if $R \geq R_U$}\\
    \left(\frac{R_U - R}{R_U - R_L}\right)^2 & \text{else}\\
\end{cases}
\end{equation}

\begin{figure}[ht!]
\centering
	\hspace*{0cm}\includegraphics[trim={0cm 0cm 0.0cm 0.0cm},clip,width=0.8\linewidth]{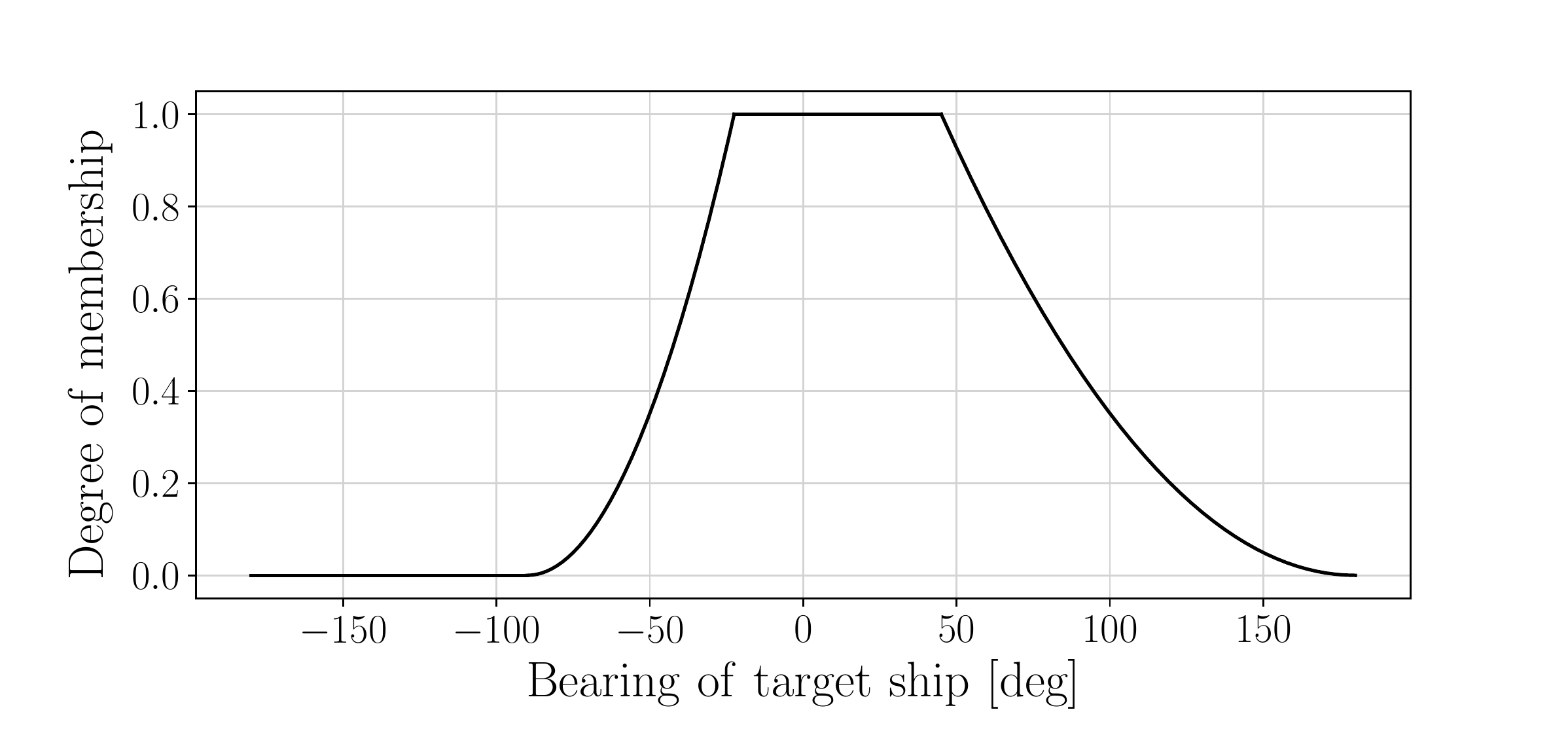}
	\caption[Bearing angle membership function]{Membership function for the bearing angle, $\theta_T$, of the target ship, with bounds $\theta_{PU}$ = 180$\degree$, $\theta_{PL}$ = 45$\degree$, $\theta_{NU}$ = 90$\degree$, and $\theta_{NL}$ = 22.5$\degree$.}
	\label{fig:mem_theta}
\end{figure}

To encourage the appropriate behavior in head-on situations, the function for the bearing angle of the TS relative to the OS should be largest on the starboard side. Defining $\theta_{PU}$, $\theta_{PL}$, $\theta_{NU}$, and $\theta_{NL}$ as the positive upper, positive lower, negative upper, and negative lower bounds on $\theta_T$, the membership function for the bearing angle can be defined as below and illustrated in \Cref{fig:mem_theta}.

\begin{equation}\label{eq:uTheta}
    u_{\theta_T} = 
    \begin{cases}
        \text{clip}\left(\left(\frac{\theta_{PU} - \theta_T}{\theta_{PU} - \theta_{PL}}\right)^2, 0, 1\right) & \text{if $\theta_T \geq 0$}\\
        \text{clip}\left(\left(\frac{\theta_{NU} - |\theta_T|}{\theta_{NU} - \theta_{NL}}\right)^2, 0, 1\right) & \text{if $\theta_T < 0$}\\
    \end{cases}
\end{equation}

After implementing a CRI containing the four membership functions introduced so far, it became clear that it was necessary to add an element to the CRI to deter the OS from crossing ahead of a TS. Since the TS's speed towards the OS can quantify whether the OS is ahead of the TS and is readily available in the observation vector ($v_y$ and $v_x$), an additional membership function is designed. Hence, we define $u_V(\cdot)$ as the ratio of the TS's speed towards the OS to its absolute speed, as described in \Cref{eq:uV}. Such a ratio was chosen to avoid issues with differences in speed among the TSs, which quickly could have arisen if the numerical value of $v_y$ had been used instead. On the other hand, it might be desirable to distinguish between crossing ahead ships traveling at different speeds, as faster ships naturally pose a higher risk. However, this is considered to be outside the scope of this work. It is worth noting that $u_V(\cdot)$ is negative when $v_y$ is negative, emphasizing the advantage of astern crossings.
\begin{equation}\label{eq:uV}
u_{V} = 
    \frac{v_y}{\sqrt{v_x^2 + v_y^2}}
\end{equation}
Integrating the introduced membership functions into a collision risk index, we have that
\begin{equation}\label{eq:CRI}
   CRI = \text{max}\left(0, \alpha_{CPA} \sqrt{u_{DCPA} \cdot u_{TCPA}} + \alpha_{\theta_T} u_{\theta_T} + \alpha_R  u_R + \alpha_V  u_V \right)
\end{equation} 

where the CPA composite term was designed in such that a combination of low values for both DCPA and TCPA gives rise to a high CRI. It also accurately expresses how a low value of either DCPA or TCPA significantly reduces the overall risk. The \texttt{max}-function is applied to ensure that the CRI is always larger or equal to zero.

Finally, values are assigned to the weights such that the sum is equal to unity, giving

\begin{equation}
  \alpha_{CPA} + \alpha_{\theta_T} + \alpha_R + \alpha_V = 1
\end{equation}

In this work, the parameter values specified in \Cref{tab:risk_params} are used. Initial choices were made based on values suggested in the literature (\cite{Chen2014, yan2002}), emphasizing DCPA and TCPA. However, it was discovered that more weight had to be placed on the target bearing angle, absolute distance, and approaching velocity to achieve the desired behavior. The configuration of the path following and static obstacle rewards listed in \cite{Meyer2020} have been applied in this work.

\begin{table*}[htb]
\caption[Reward configuration for the risk-based approach]{Reward configuration for the risk-based approach.}\label{tab:risk_params}
    \centering
	\begin{tabular}{lll}
		\hline
		Parameter & Interpretation & Value\\
		\hline
		$\beta_{CRI}$ & Scaling factor for overall risk level & $10$ \\
		$\beta_{RL}$ & Scaling factor for the lower bound on distance & $8$ \\
		$\beta_{RU}$ &  Scaling factor for the upper bound on distance & $18$ \\
		$\theta_{PU}$ &  Positive upper limit for the bearing angle $\theta_T$ & $180\degree$ \\
		$\theta_{PL}$ &  Positive lower limit for the bearing angle $\theta_T$ & $45\degree$ \\
		$\theta_{NU}$ &  Negative upper limit for the bearing angle $\theta_T$ & $90\degree$ \\
		$\theta_{NL}$ &  Negative lower limit for the bearing angle $\theta_T$ & $22.5\degree$ \\
		$\alpha_{CPA}$ & Weighting of CPA membership function & $0.3$ \\
		$\alpha_{\theta_T}$ & Weighting of target bearing angle membership function & $0.2$ \\
		$\alpha_R$ & Weighting of absolute distance membership function & $0.3$ \\
		$\alpha_V$ & Weighting of approaching velocity membership function & $0.2$ \\
		$d_L$ & Minimal safe encounter distance & $320$ m \\
		$d_U$ & Absolute safe encounter distance & $1500$ m \\
		\hline
	\end{tabular}
\end{table*}

\subsection{Performance evaluation}
A three-step evaluation process is employed to assess the performance of the RL agent. First, the agent's behavior and performance in the training environment are assessed, and snippets from situations relevant to rules 14-16 of the COLREGs are presented. Next, two-vessel testing scenarios are constructed to test for COLREG-compliance specifically. Lastly, the agents are evaluated in AIS-based environments. These modes of assessment are described individually in the following subsections.

\subsubsection{Performance in the training environment}
A natural starting point for performance evaluation is assessing the agent's behavior in its training environment. The overall performance can be evaluated by collecting statistics on the collision rate, level of path completion, and reward. These statistics serve as a guide for when to stop the training and a point of comparison between approaches. Moreover, a qualitative assessment is made by observing the agent's behavior through video recordings. Snippets are chosen from the videos to highlight the behavior in situations where the COLREGs apply. This is not always the case since the training environment often presents the agent with difficult situations containing various static and dynamic obstacles, which cannot be accurately subjected to the COLREGs.

\subsubsection{Testing of COLREG-compliance}
The next step in the testing process is subjecting the agent to scenarios specifically designed to capture COLREG-compliance. This is especially useful since it is challenging to find scenarios that perfectly showcase COLREG-compliance in the training environment. However, the agent's success can easily be quantified through simpler two-vessel scenarios. One scenario to be tested is self-evident, namely the head-on scenario. In addition, two different crossing situations, one from the starboard and one from the port side, were chosen. For each scenario, the TS's initial angles and path angles are varied slightly within a range of $\pm 5\degree$ of the default angles, which allows for an accumulation of statistics on success rate in the respective scenarios. It should be noted that the target ships have been modeled exclusively large in the testing scenarios to reflect the size of the large ships encountered in the AIS-based scenarios and for visual clarity. 

\subsubsection{AIS-based testing}
Lastly, three environments based on real-world high-fidelity terrain data are used to assess the generalization performance of the agent. These environments were developed by \cite{Meyer_thesis} using AIS tracking data and terrain data from the Trondheim Fjord area and are distinctly different.  A dashed black line represents the desired OS trajectory in the following illustrations. Each TS is drawn at its initial position, and trajectories are drawn as dotted red lines. Note that these are examples of spawned environments and that a set amount of target ships are chosen from the AIS database each time an instance of the specific scenario is created. Additionally, the apparent density of TS trajectories does not directly reflect the number of encounters, as this depends on the speed of each vessel. 

The first AIS-based scenario is the \textbf{Trondheim} scenario (\Cref{fig:trondheim_testscenario_res_colregs_risk}), in which the agent is required to cross a fjord of width $\sim$12 km while following a straight path. Doing so, it mainly meets crossing traffic consisting of larger vessels. In the challenging \textbf{Ørland-Agdenes} scenario (\Cref{fig:orlandagdenest_testscenario_res_colregs_risk}), the agent encounters two-way traffic in a narrow fjord entrance region. It must blend into the heavy traffic to complete the path while avoiding head-on collisions. In addition, the ability to overtake other vessels is assessed. As in the Trondheim scenario, the vessels are primarily bigger than the OS. Lastly, the \textbf{Froan} scenario (\Cref{fig:sorbuoyatestscenario_res_colregs_risk}) offers a demanding terrain with hundreds of small islands. As a result, it tests the ability of the agent to generalize to a challenging environment with a high density of static obstacles in varying shapes and sizes. The area is less trafficked, and the vessels encountered are physically similar to the OS.

\section{Results and discussion}
\label{sec:results}
In this section, the results from the risk-based implementation of the COLREGs are presented and evaluated. First, the RL agent is evaluated in the synthetic training environment, considering its general path following and collision avoidance performance. Second, the presence and consistency of COLREG-compliant behavior are assessed in isolated, high-risk encounters. Finally, the agent is presented the simulated real-world AIS-based scenarios to see how the learned policy generalizes to complex and unseen situations.

\subsection{Training and testing in the synthetic environment}\label{sec:results_risk_based}
\begin{figure}[htb!]
    \captionsetup[subfigure]{justification=centering}
	\begin{center}
	    \subfloat[Head-on situation]{\includegraphics[width=0.45\linewidth]{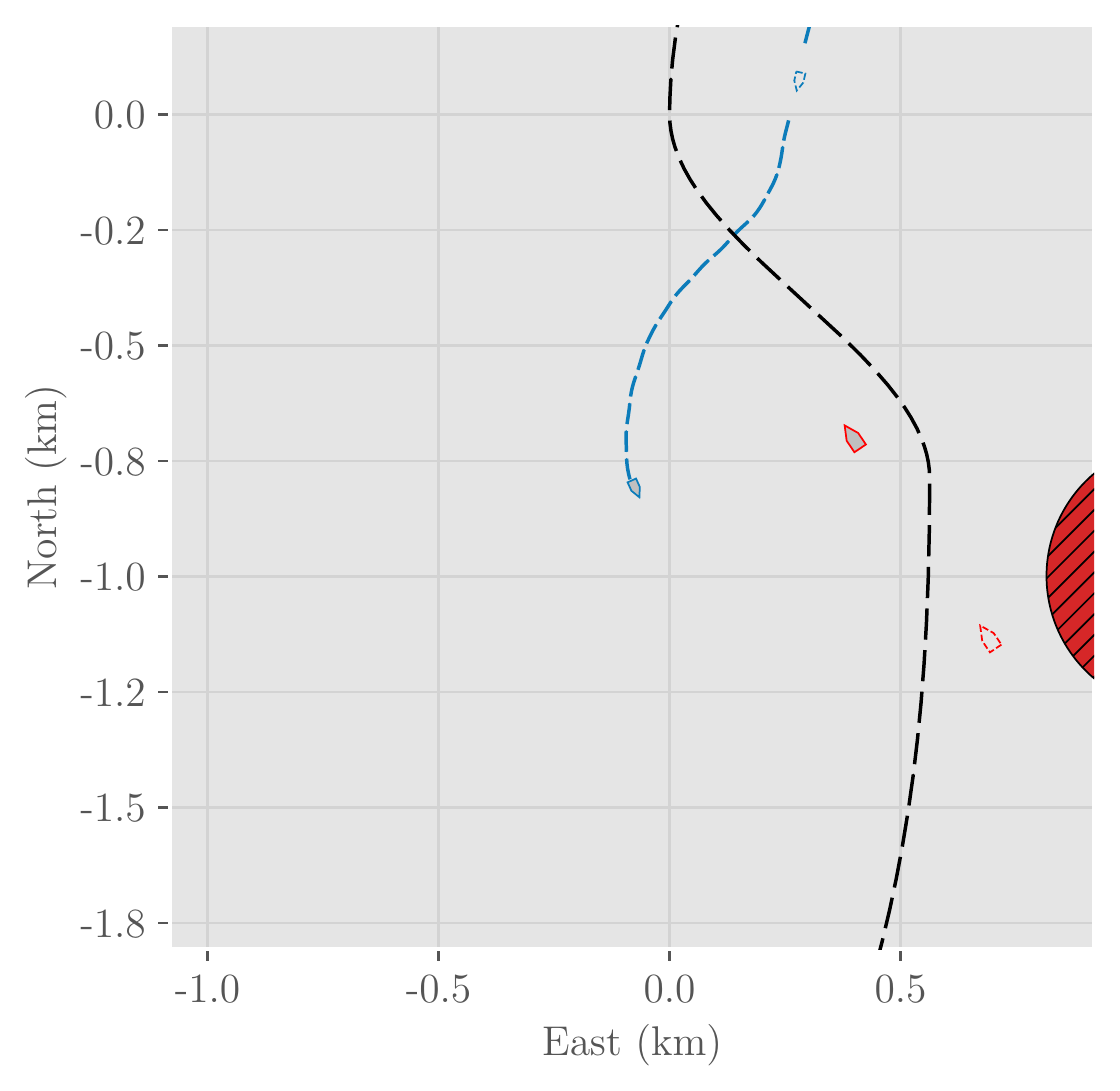}\hspace{-1pt}%
	    }
	    \subfloat[Astern passing]{\includegraphics[width=0.45\linewidth]{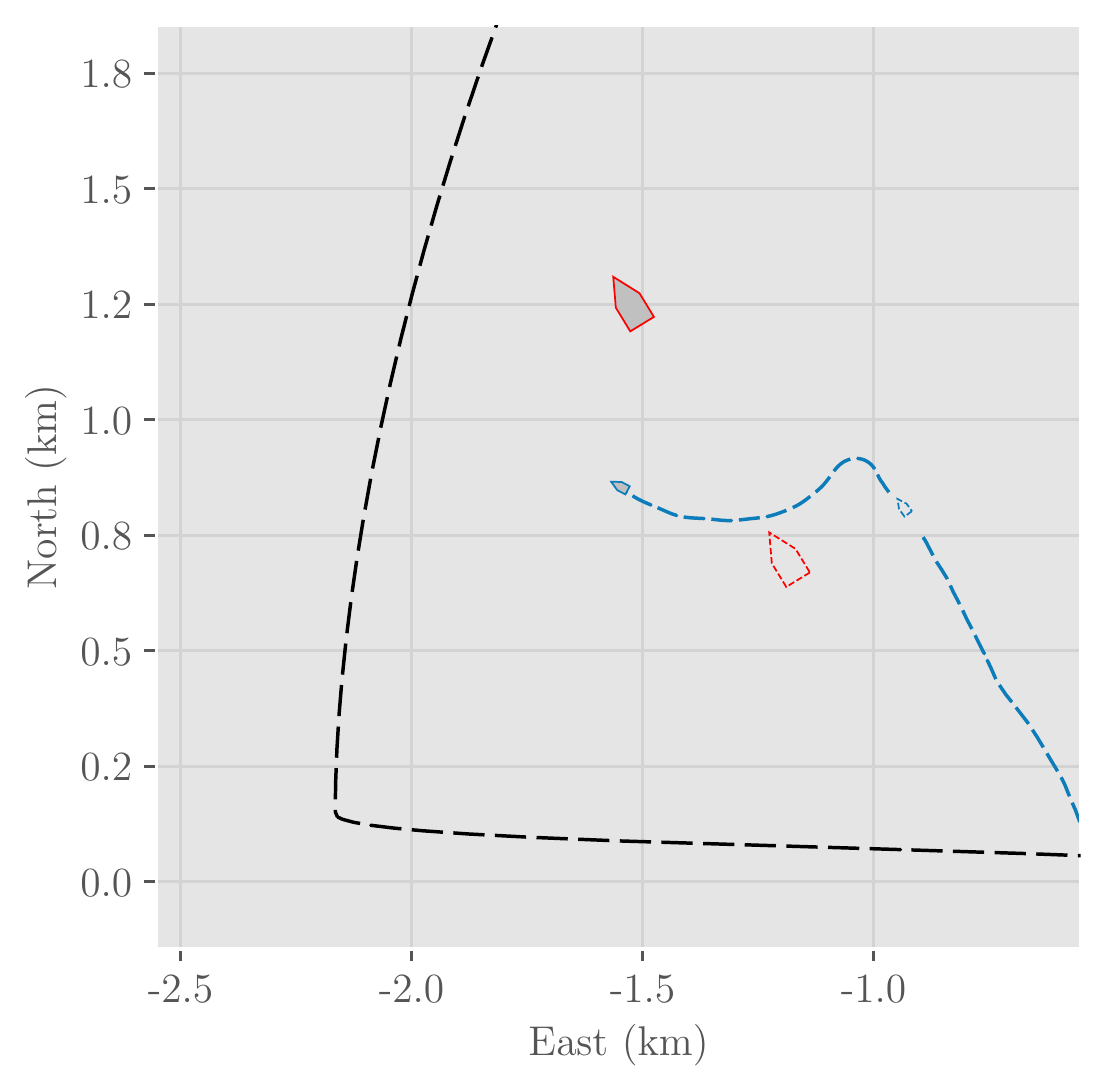}\hspace{1pt}%
	    }
	\end{center}
	\begin{center}
	    \subfloat[Overtaking]{\includegraphics[width=0.45\linewidth]{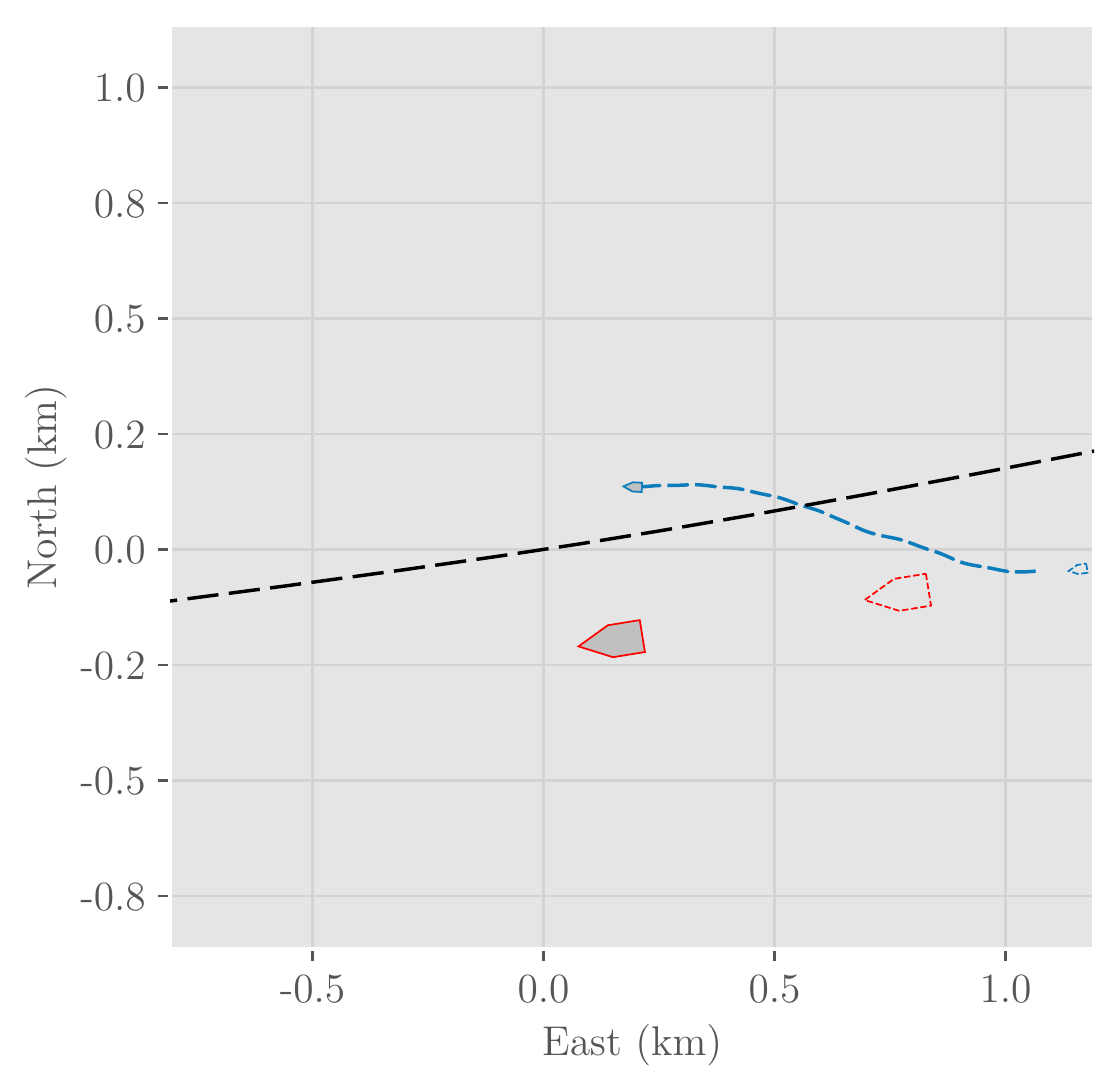}\hspace{-1pt}%
  	    }
  	    \subfloat[Static COLAV]{\includegraphics[width=0.45\linewidth]{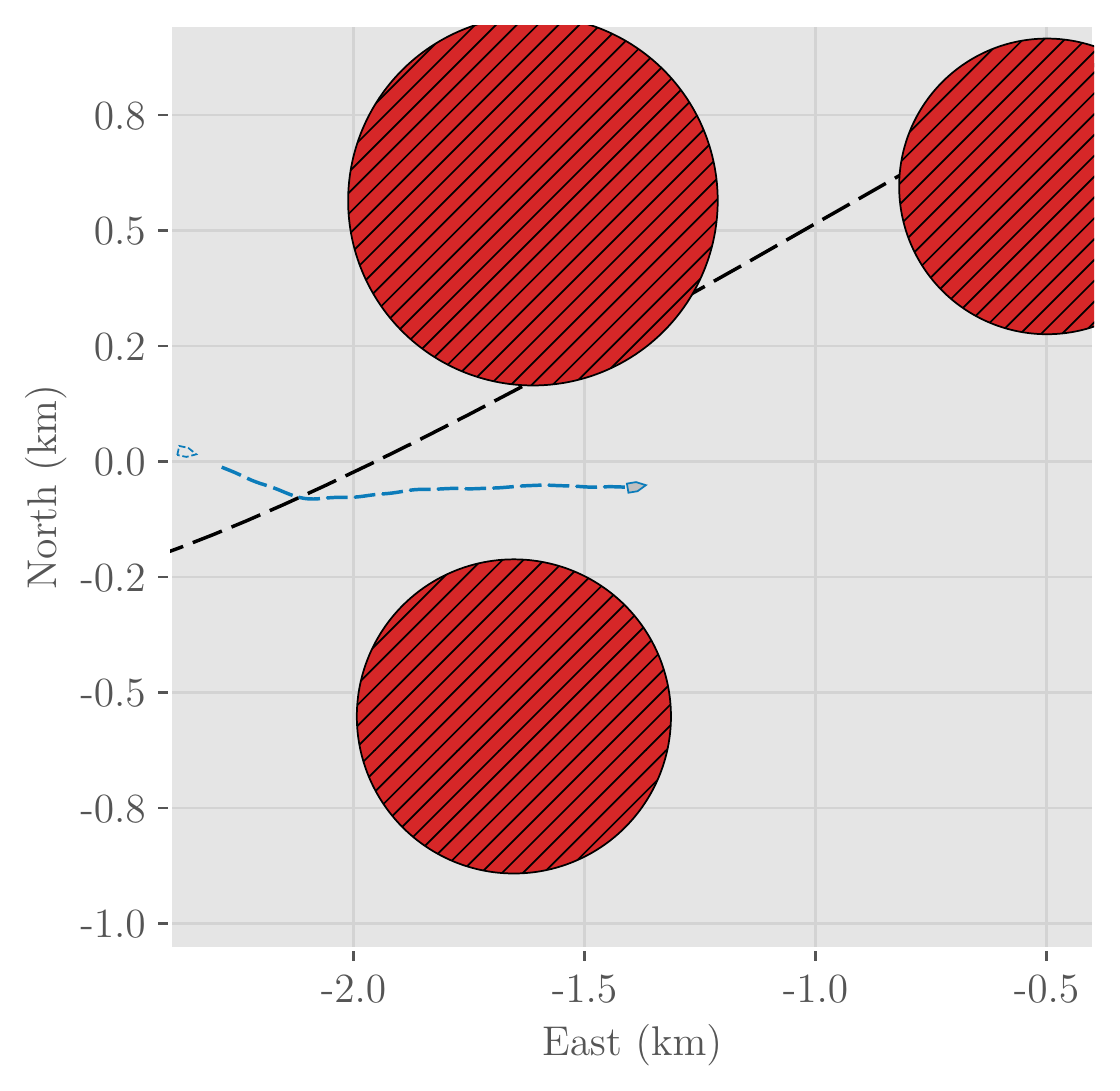}\hspace{1pt}%
  	    }
	\end{center} 	 
	\caption[Risk-based approach: snippets from training environment]{Risk-based agent performing common naval collision avoidance maneuvers in the training environment. The agent's trajectory is drawn as a blue dashed line, and the target ships with trajectories are drawn in red. The dotted vessel outlines show their positions 100 time steps prior.}
	\label{fig:training_scenario_sims_risk}
\end{figure}

After training the RL agent in the synthetic environment (\Cref{fig:movingobstacles_demo}) for approximately 4000 episodes, its collision rate dropped to near zero, and the progress rate rose to 100\%. Snippets from the training environment have been included in \Cref{fig:training_scenario_sims_risk}, showcasing training scenarios in which the agent behaves in a COLREG-compliant manner. The COLREGs clearly define these situations: passing on the right in head-on situations, slowing down and passing astern instead of ahead, and allowing space between it and the TS during overtaking. Although the training statistics indicate the agent's ability to navigate and avoid collisions, they do not reveal whether the COLREG-compliance is consistent, which must be evaluated separately.

\subsubsection{Testing of COLREG-compliance}
\begin{figure}[!htb]
    \centering
    \hspace*{-0.3cm}
    \begin{subfigure}{0.4\linewidth}
    	\centering	\hspace*{-1.0cm}
    	\hspace*{-0.3cm}\begin{overpic}[trim={0cm 0cm 0 0cm},clip, width=1.1\textwidth]{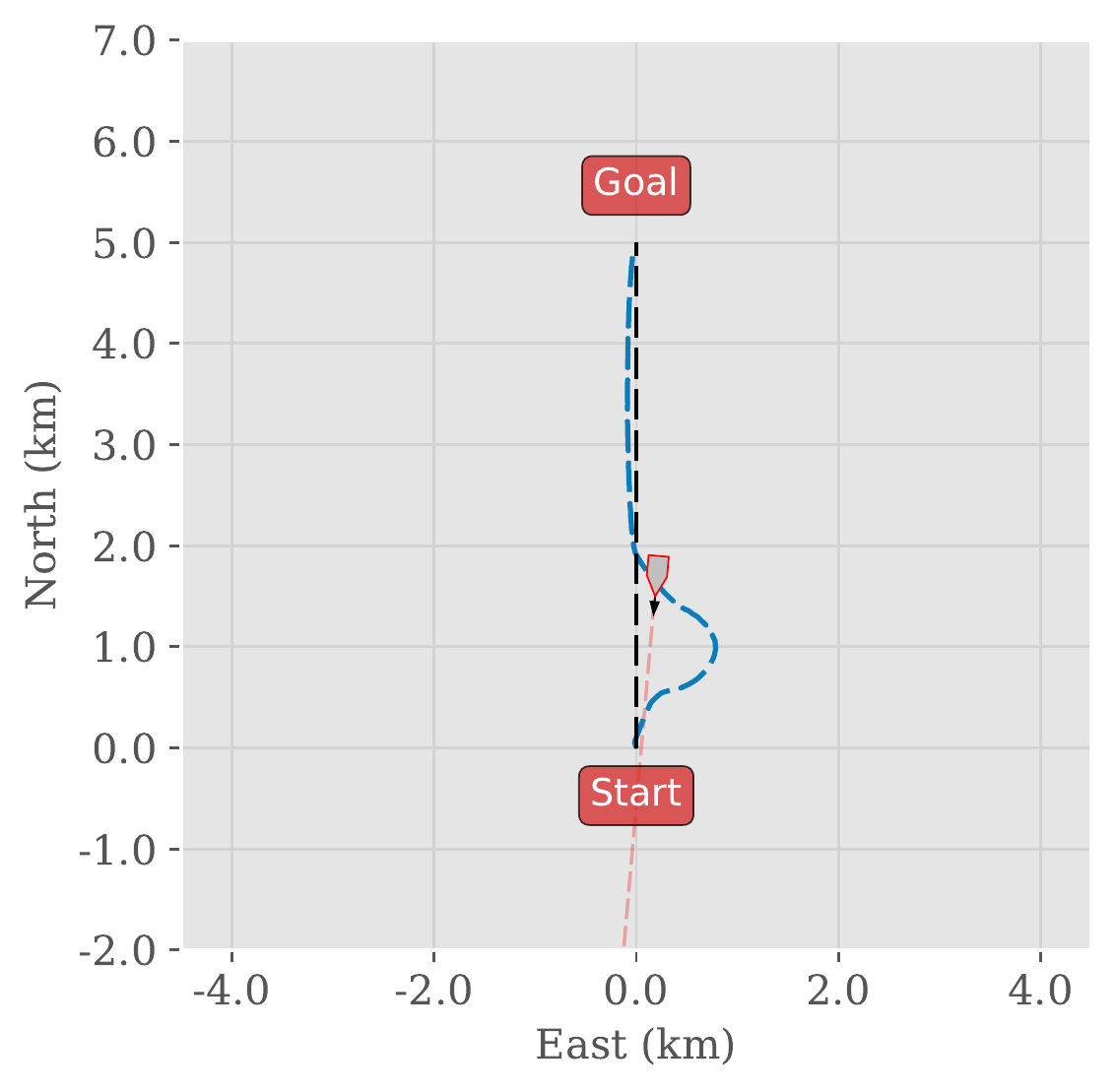}
         \put(75,50){\frame{\includegraphics[trim={4.1cm 3.5cm 4.1cm 2.5cm},clip, width=.276\textwidth, height=.42\textwidth]{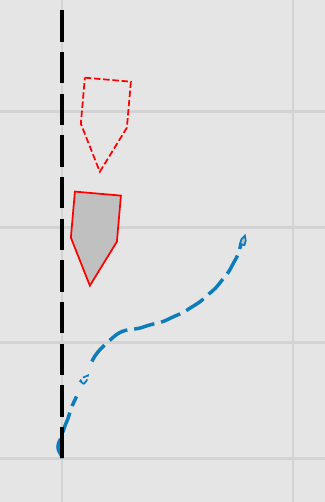}}}  
        \end{overpic}
        \subcaption[Head on test scenario]{Test scenario 1: \\Head on.}
        \label{fig:headon_testscenario_risk}
    \end{subfigure}
    \hspace*{0.8cm}
    \begin{subfigure}{0.4\linewidth}
    	\centering
    	\hspace*{-1.0cm}\begin{overpic}[trim={0cm 0cm 0 0cm},clip, width=1.1\textwidth]{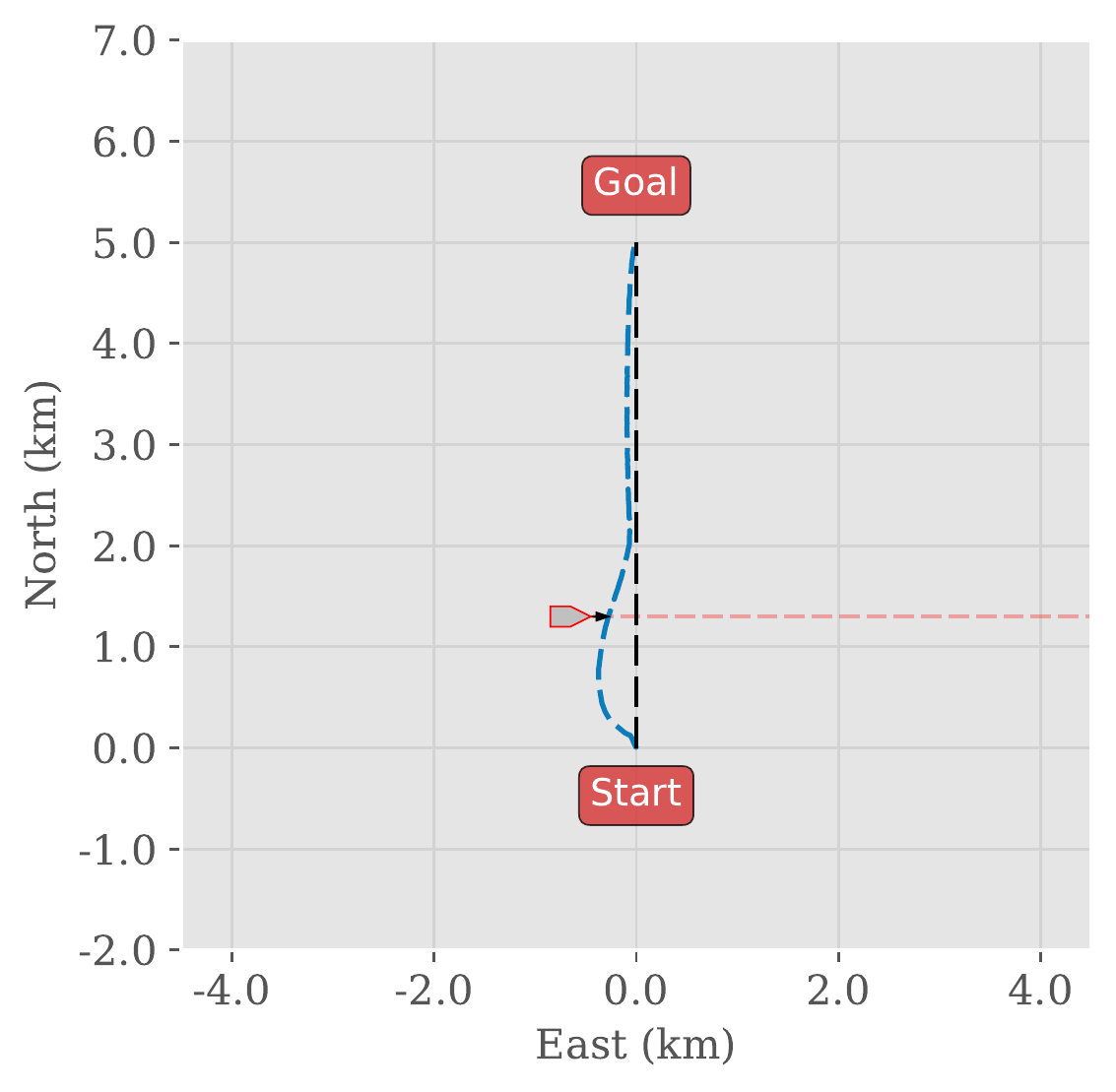}
         \put(75,50){\frame{\includegraphics[trim={5.6cm 2cm 1.5cm 3.5cm},clip, width=.276\textwidth, height=.42\textwidth]{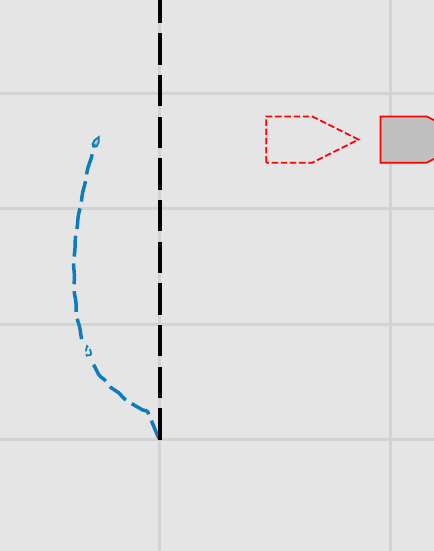}}}  
        \end{overpic}
        \subcaption[Crossing1 test scenario]{Test scenario 2:\\ Crossing from starboard.}
        \label{fig:crossing1_testscenario_risk}
    \end{subfigure}
    \begin{subfigure}{0.4\linewidth}
    	\centering
    	\vspace*{1.0cm}
    	\hspace*{-1.0cm}\begin{overpic}[trim={0cm 0cm 0 0cm},clip, width=1.1\textwidth]{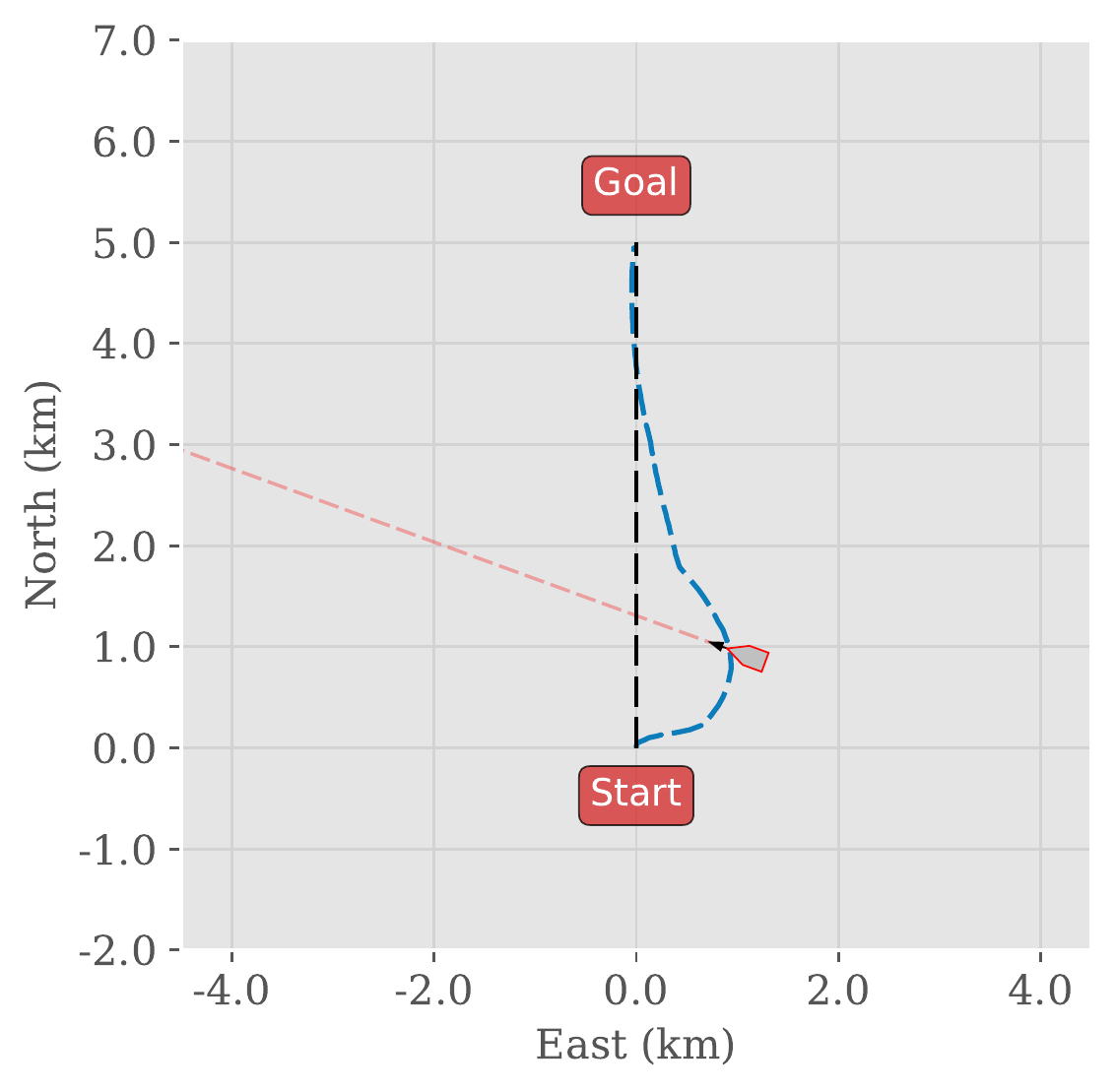}
         \put(75,50){\frame{\includegraphics[trim={4cm 3cm 4.2cm 3.6cm},clip, width=.276\textwidth, height=.42\textwidth]{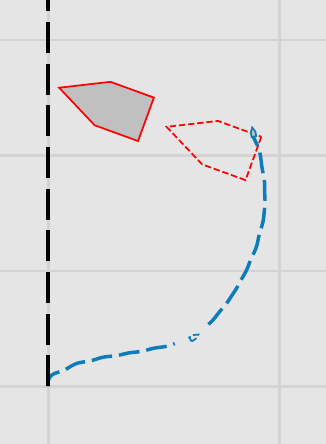}}}
         \end{overpic}
        \subcaption[Crossing2 test scenario]{Test scenario 2: \\Crossing from port.}
        \label{fig:crossing2_testscenario_risk}
    \end{subfigure}
    \caption[Risk-based approach: COLREG-compliance tests]{Agent behaviour in COLREG-compliance test scenarios. The agent's trajectory is drawn as a blue dashed line, and the target ships with trajectories are drawn in red. The dotted vessel outlines show their positions 100 time steps prior to the present time.}
    \label{fig:colregs_compliance_sims_risk}
\end{figure}

The next step in the evaluation process is COLREG-compliance testing with repetitive testing in different encounter scenarios. \Cref{fig:colregs_compliance_sims_risk} shows how the agent avoids collision in a COLREG-compliant manner. In addition, the agent follows the path well once the encounter has passed. Repetitive testing reveals that these results are stable, as the correct behavior was seen in 100\% of the episodes for each testing scenario, as summarised in \Cref{tab:risk_based_rep_tests}. These results indicate that the agent can intelligently interpolate between path following and COLREG-compliant collision avoidance in isolated high-risk encounters. However, there is no guarantee that this behavior translates into more complex scenarios.

\begin{table}[htb]
    \centering
    \caption[Risk-based approach: repetitive COLREG-compliance test results]{Results from repetitive testing of COLREG-compliance with slightly varying scenarios, 100 episodes.}\label{tab:risk_based_rep_tests}
	\begin{tabular}{ll}
		\hline
		Scenario & Success rate\\
		\hline
		Head-on & 100\% \\
		Crossing from starboard & 100\% \\
		Crossing from port & 100\% \\
		\hline
	\end{tabular}
\end{table}

\subsection{Testing in AIS-based environments}

\begin{figure}[!htb]
    \captionsetup[subfigure]{justification=centering}
	\begin{center}
	    \subfloat[Head-on situation]{\includegraphics[width=0.45\linewidth]{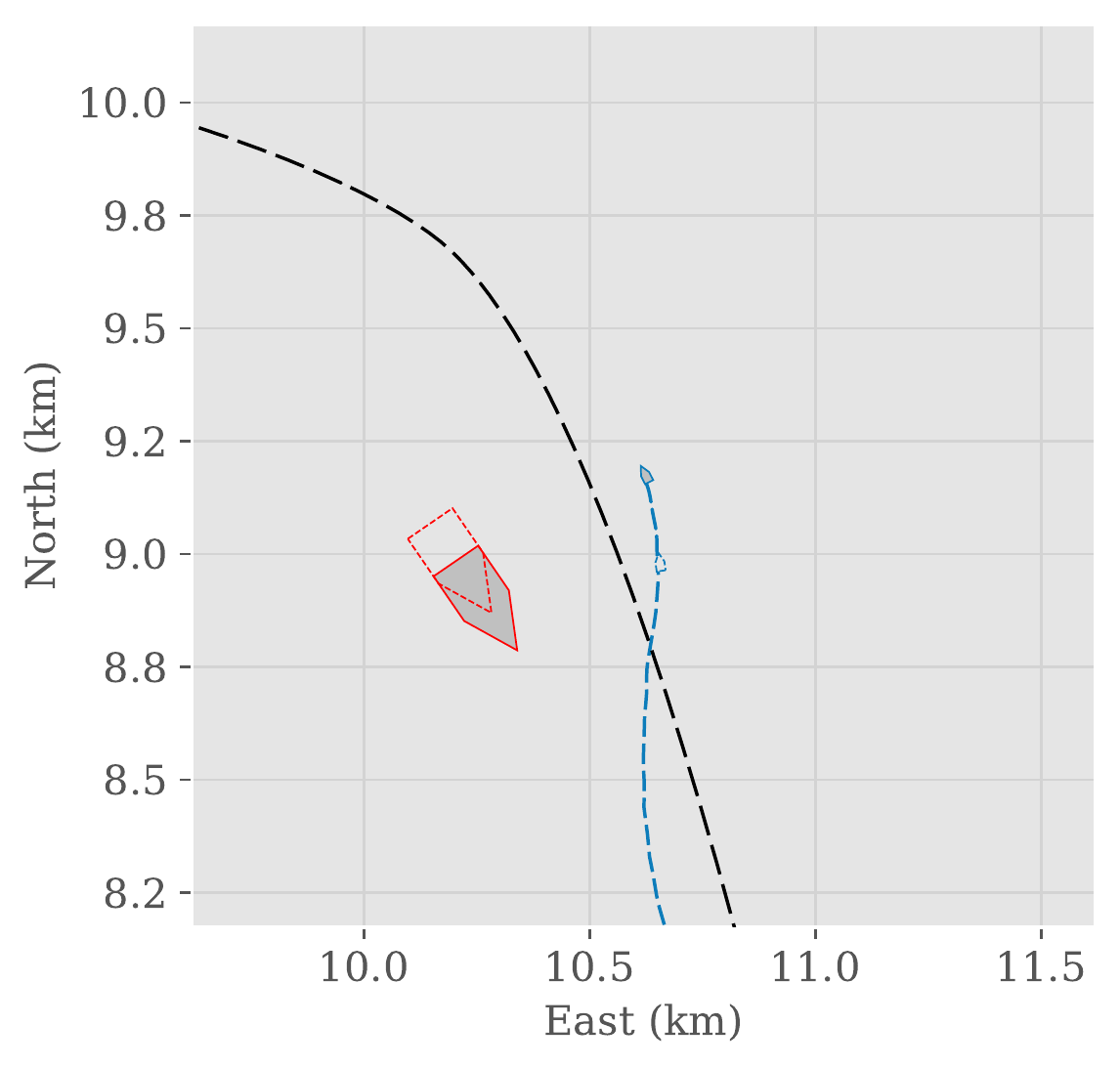}\hspace{-1pt}%
	    }
	    \subfloat[Astern passing]{\label{fig:astern_ais_risk}\includegraphics[width=0.45\linewidth]{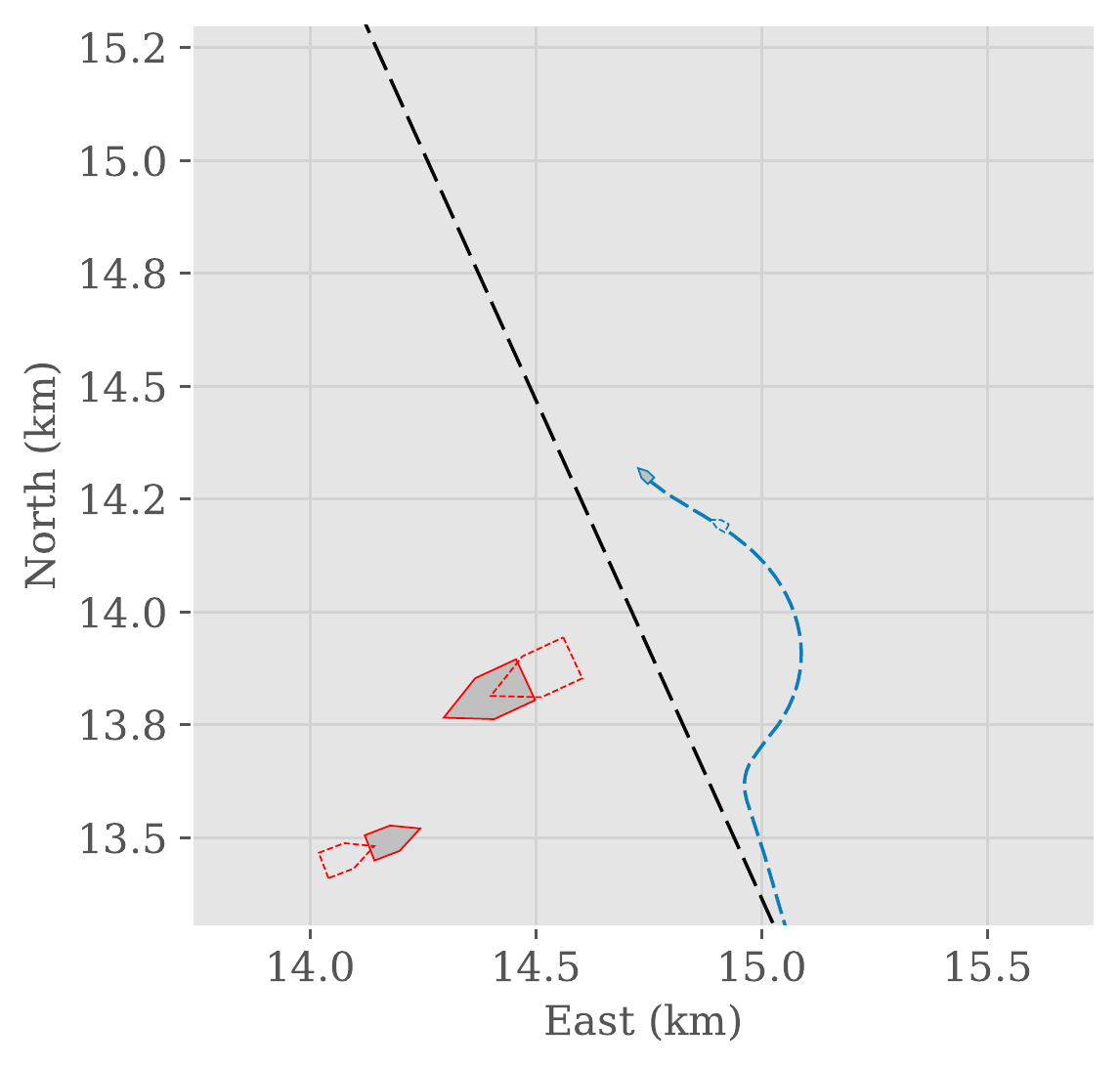}\hspace{1pt}%
	    }
	\end{center}
	\begin{center}
	    \subfloat[Overtaking]{\includegraphics[width=0.45\linewidth]{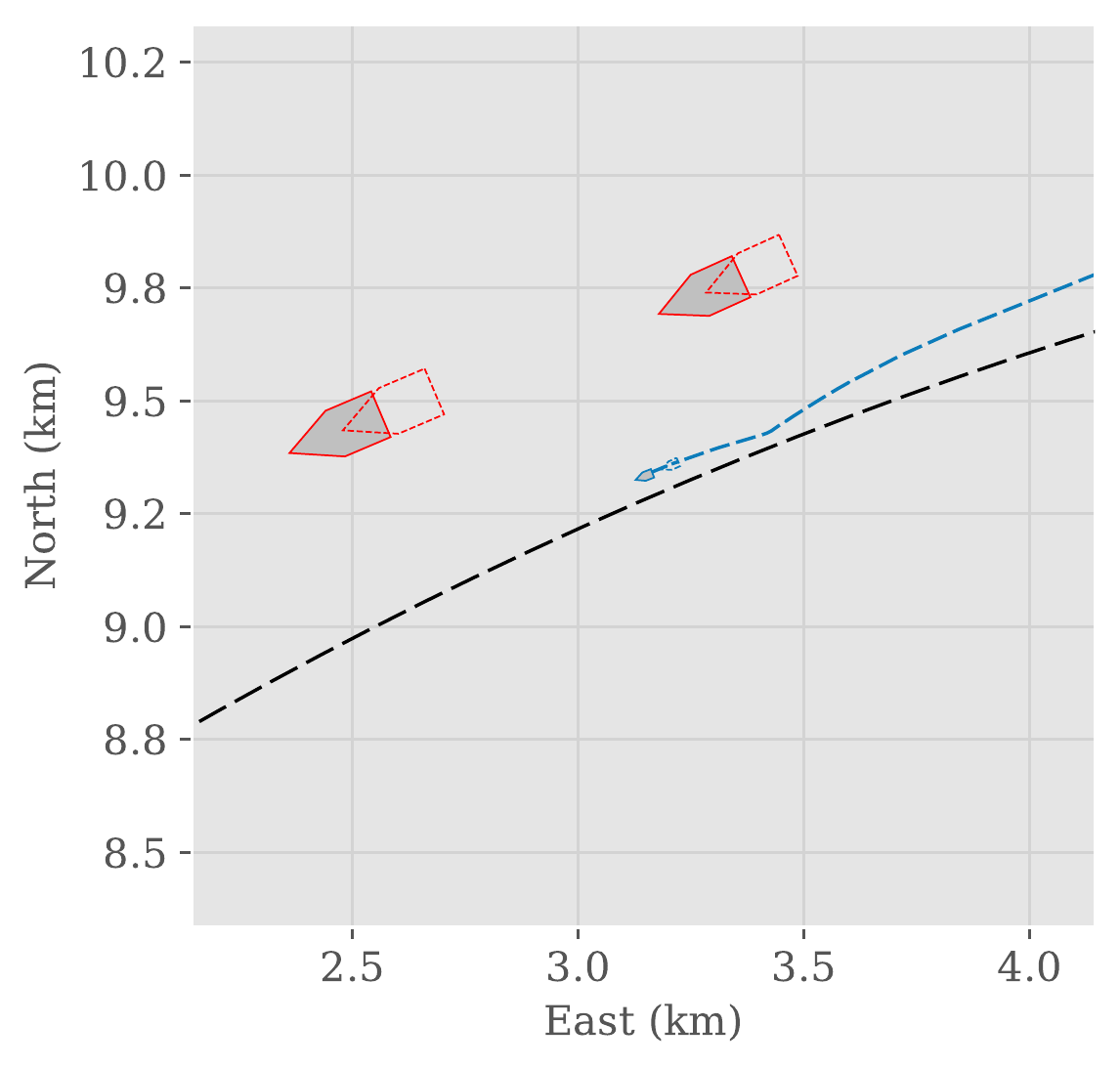}\hspace{-1pt}%
  	    }
  	    \subfloat[Static COLAV \label{fig:RW_static_risk}]{\includegraphics[width=0.45\linewidth]{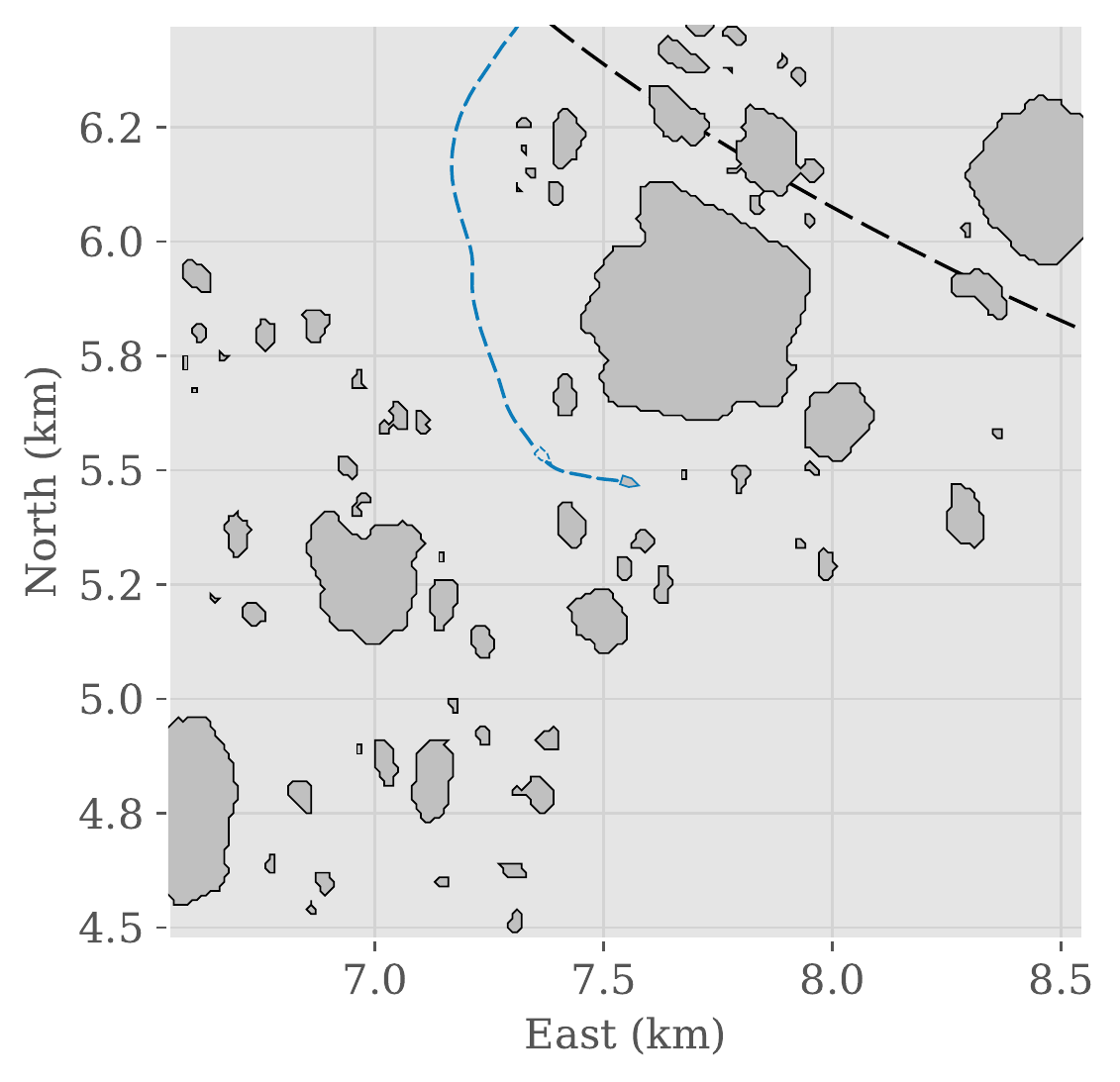}\hspace{1pt}%
  	    }
	\end{center} 	 
	\caption[Risk-based approach: snippets from AIS-based environment]{Risk-based agent performing common naval collision avoidance maneuvers in the AIS-based environment. The agent trajectory is drawn as a blue dashed line, and the target ships are drawn in red. The dotted vessel outlines show their positions 100 time steps prior to the present time.}
	\label{fig:real_scenarios_sims_colregs_risk_based}
\end{figure}

Finally, the risk-based agent is assessed in AIS-based real-world environments to find how well the agent generalizes to previously unseen scenarios. As these environments are modeled using real-world terrain mapping and AIS traffic data, the agent will likely encounter complex scenarios where the COLREGs do not clearly define the correct behavior. Therefore, we do not expect the agent always to find a COLREG-compliant solution. The agent's excellent static obstacle avoidance and COLREG-compliant behavior are highlighted in \Cref{fig:real_scenarios_sims_colregs_risk_based}. Note that the static obstacles in \Cref{fig:RW_static_risk} are significantly smaller than those encountered in the training scenario. 

\begin{figure}[!htb]
	\centering
	\begin{subfigure}{0.32\linewidth}
		\centering
		\includegraphics[width=\textwidth]{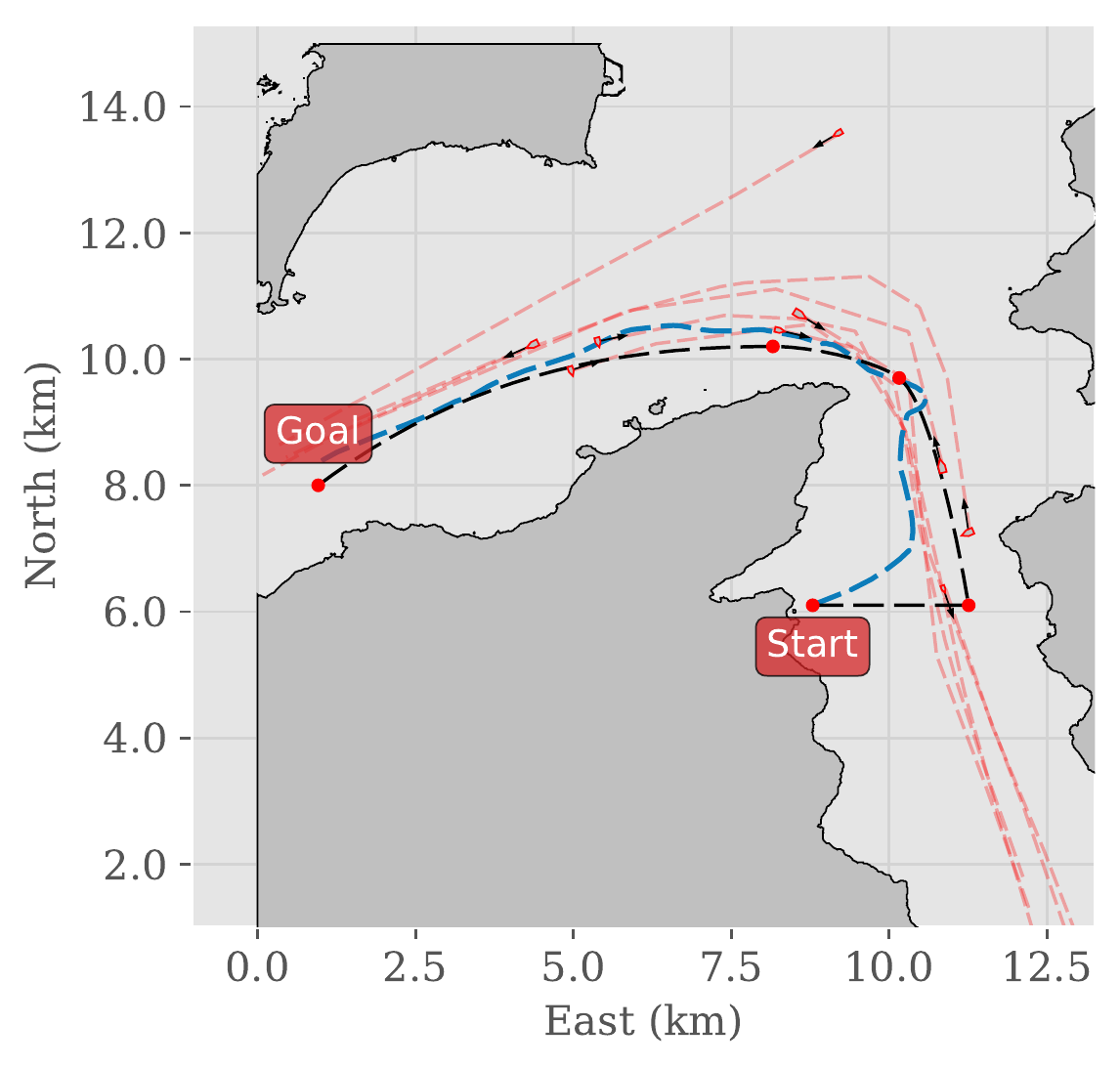}
	    \subcaption[Ørland-Agdenes test scenario result]{Risk-based agent's trajectory in the Ørland-Agdenes test scenario. The agent intelligently merges with the two-way traffic, and follows it towards the goal.}
	    \label{fig:orlandagdenest_testscenario_res_colregs_risk}
	\end{subfigure}
	\begin{subfigure}{0.32\linewidth}
		\centering
		\vspace*{0.3cm}\includegraphics[width=\textwidth]{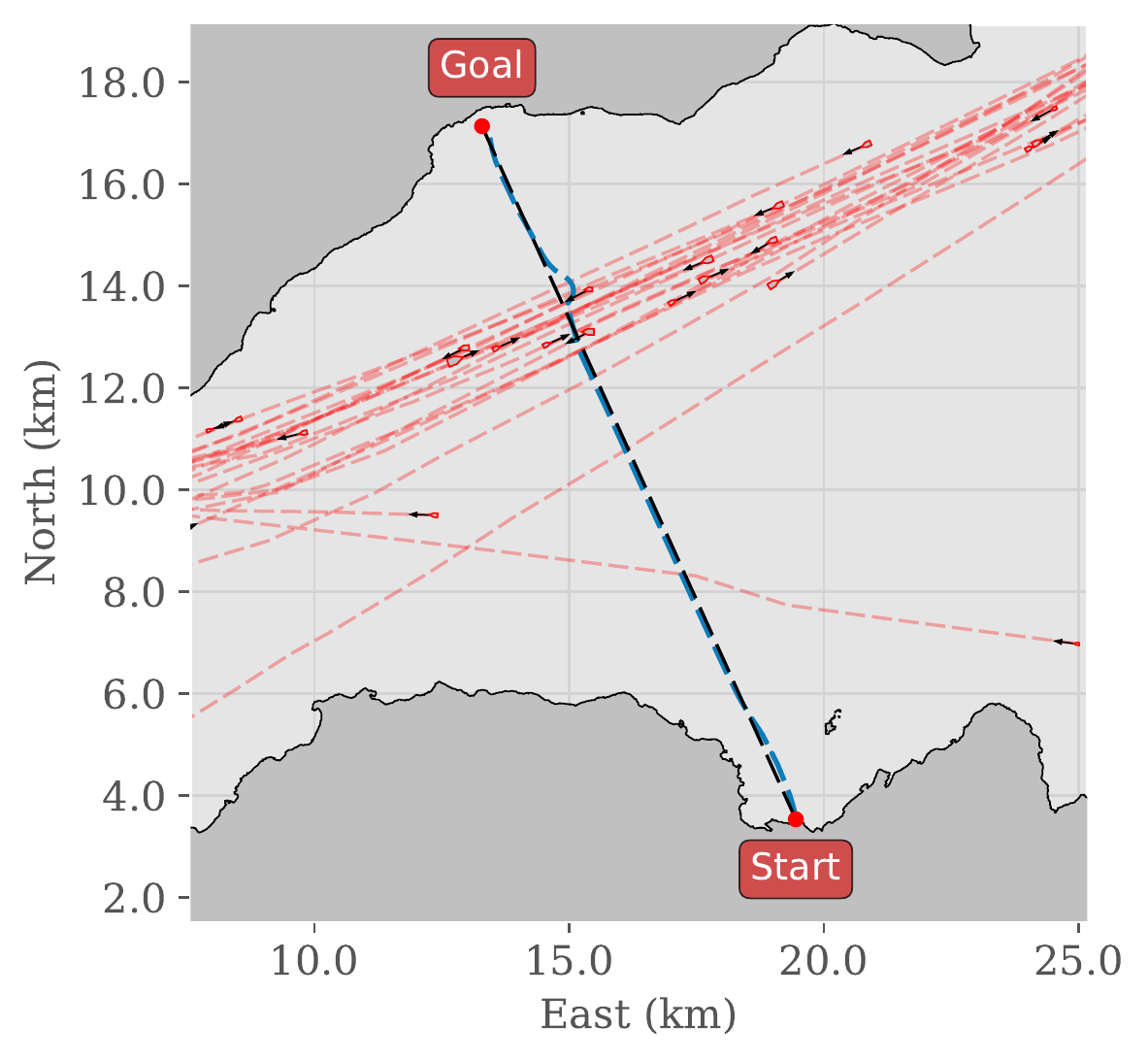}
	    \subcaption[Trondheim test scenario result]{Risk-based agent's trajectory in the Trondheim test scenario. The agent exhibits excellent path following, and is seen to maneuver the crossing traffic before returning to the path and reaching the goal.}
	    \label{fig:trondheim_testscenario_res_colregs_risk}
	\end{subfigure}
	\begin{subfigure}{0.32\linewidth}
		\centering
		\vspace*{0.5cm}\includegraphics[width=\textwidth]{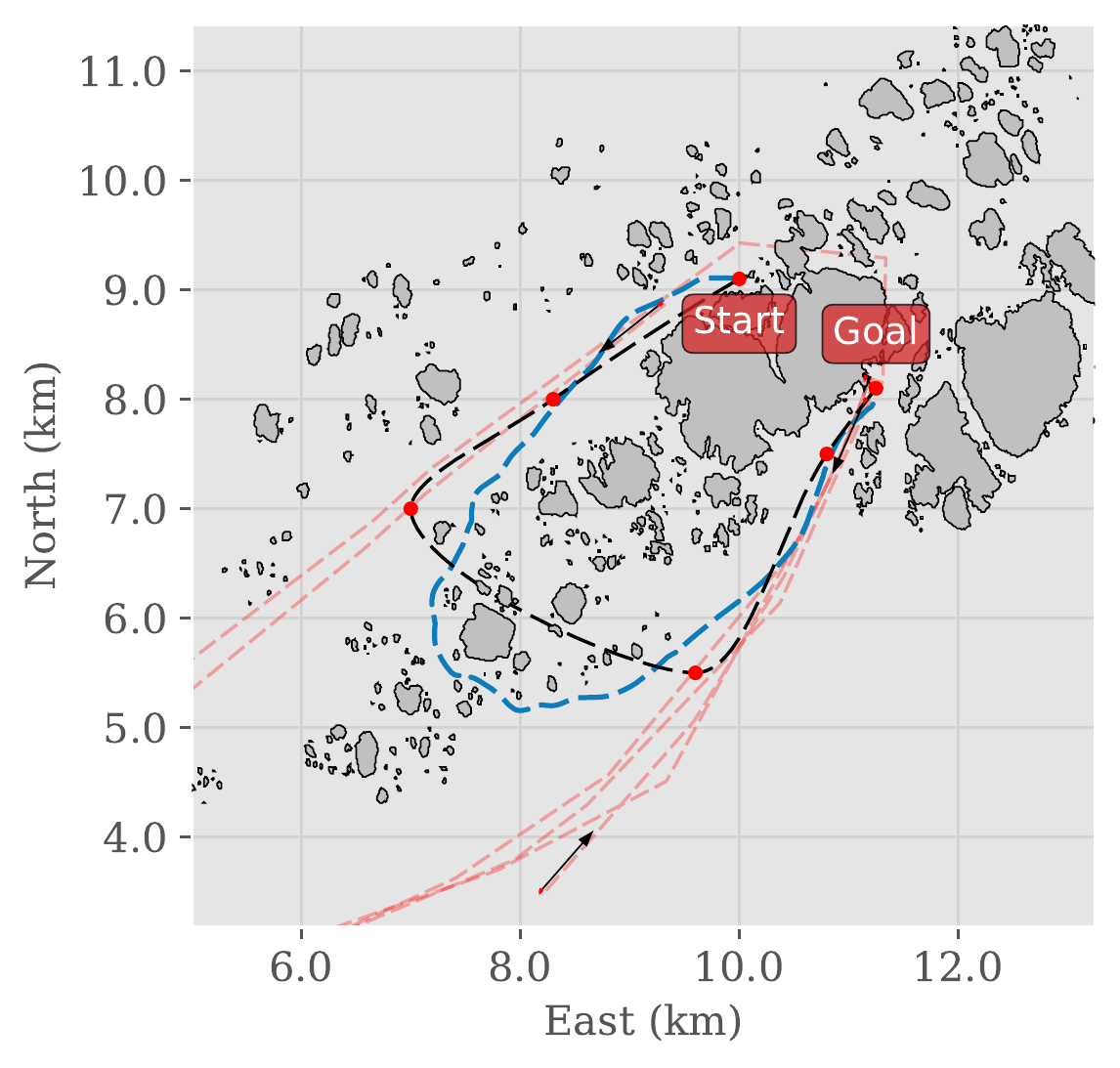}
	    \subcaption[Froan test scenario result]{Risk-based agent's trajectory in the Froan test scenario. When presented with an impossible path, the agent intelligently finds another solution for navigating the dense archipelago and merge into the parallel traffic.}
	    \label{fig:sorbuoyatestscenario_res_colregs_risk}
	\end{subfigure} 
	\caption{Trajectories from three different AIS-based environments drawn as blue dashed lines. Target ships and trajectories are drawn in red.}
	\label{fig:real_world_sims_colregs_risk_based}
\end{figure}

Lastly, trajectories from each environment are presented in \Cref{fig:real_world_sims_colregs_risk_based}. These trajectories illustrate the agent's ability to dynamically follow a predetermined path in the face of static and moving obstacles. Whether the agent is faced with heavy two-way parallel traffic (\Cref{fig:orlandagdenest_testscenario_res_colregs_risk}), crossing traffic (\Cref{fig:trondheim_testscenario_res_colregs_risk}), or an untraversable path (\Cref{fig:sorbuoyatestscenario_res_colregs_risk}), it adapts to the situation and finds a suitable solution. Thus, the agent generalizes its decision-making policy from the synthetic and stochastic training environment to previously unseen environments.


\section{Conclusion}
\label{sec:conclusion}

The primary objective of this work was to investigate whether COLREG-compliance in a path following and collision avoidance system based on model-free DRL is possible using a risk-based approach. In summary, we found that
\begin{itemize}
    \item using state-of-the-art collision risk theory, the ambiguity of the COLREGs (rules 14-16) can be circumvented by conditioning the decision-making agent on collision risk indices as a proxy for the COLREGs. 
    \item the agent learned complex rules by training in a stochastic and synthetic training environment, which translated well into real-world testing environments. Moreover, the approach produced COLREG-compliant behavior when tested in isolated encounters.
\end{itemize}

As described in \Cref{sec:methods:perception}, the agent perceives its surroundings by summarizing the information from N distance sensors into D sectors using feasibility pooling. Consequently, the agent directly reads the decomposed velocity of the nearest TS in each sector to ensure their detection. This approach leads to significant information loss, as the agent can neither observe high-frequency details, such as the size of the TS, nor if there are multiple obstacles in a single sector. The curse of dimensionality, which motivates the feasibility pooling algorithm, can potentially be avoided by applying a convolutional neural network (CNN). Unlike fully connected networks, CNNs directly utilize spatial information in structured data. Thus, a CNN can exploit the fact that there is a strong correlation between neighboring sensor measurements, given that the resolution is high enough. Furthermore, by stacking multiple observations temporally, the observation then contains sufficient information for the agent to infer any obstacle's relative velocity and acceleration from the distance measurements alone, removing the need for object tracking and velocity estimation methods.

Additionally, the COLREGs must be adapted for machine readability and interpretability, clearly defining the required behavior in different environments and situations. Without doing so, it will be impossible to accurately assess the success of an autonomous vessel and claim that it is fully COLREG-compliant. Nevertheless, the results obtained in this work imply the potential of DRL to handle COLREG rules through risk-based conditioning. Thus, once the COLREGs are modernized for digital applications, DRL can be expected to produce COLREG-compliant and autonomous collision avoidance systems.

\section*{Acknowledgments}
The authors acknowledge the financial support from the Norwegian Research Council and the industrial partners: DNV GL, Kongsberg and Maritime Robotics of the Autosit project. (Grant No.: 295033).
  \bibliographystyle{elsarticle-harv} 
  \bibliography{references.bib}
\end{document}